%% file: templatePRIME.tex
\documentclass{article}

\usepackage{PRIMEarxiv}

\usepackage[utf8]{inputenc} 
\usepackage[T1]{fontenc}    
\usepackage{hyperref}       
\usepackage{url}            
\usepackage{booktabs}       
\usepackage{amsfonts}       
\usepackage{nicefrac}       
\usepackage{graphicx}
\usepackage{booktabs}
\usepackage{lscape}

\usepackage{amsmath}
\usepackage{lscape}

\newcommand{\paratitle}[1]{\vspace{1.5ex}\noindent \textbf{#1}}

\title{Advances and Challenges in Deep Lip Reading
\thanks{\textit{This paper has been submitted to Computer Vision and Image Understanding(manuscript number:CVIU-21-732).}} 
}

\author{
  Marzieh Oghbaie\thanks{Corresponding Author}\\
  Imam Khomeini International University, Qazvin, Iran \\
  \texttt{marzieh.oghbaie@gmail.com} \\
   \And
  Arian Sabaghi\\
  Amirkabir University of Technology(Tehran Polytechnic), Tehran, Iran\\
  \texttt{arian.s@aut.ac.ir} \\
\And
  Kooshan Hashemifard  \\
  University of Alicante, Alicante, Spain\\
  \texttt{k.hashemifard@ua.es} \\
\And
  Mohammad Akbari  \\
  Amirkabir University of Technology(Tehran Polytechnic), Tehran, Iran\\
  School of Computer Science, Institute for Research in Fundamental Sciences (IPM), Tehran, Iran\\
  \texttt{akbari.ma@aut.ac.ir} \\

}

\begin{document}
\maketitle

\begin{abstract}
Driven by deep learning techniques and large-scale datasets, recent years have witnessed a paradigm shift in automatic lip reading. While the main thrust of Visual Speech Recognition (VSR) was improving accuracy of Audio Speech Recognition systems, other potential applications, such as biometric identification, and the promised gains of VSR systems, have motivated extensive efforts on developing the lip reading technology.
This paper provides a comprehensive survey of the state-of-the-art deep learning based VSR research with a focus on data challenges, task-specific complications, and the corresponding solutions. Advancements in these directions will expedite the transformation of silent speech interface from theory to practice. We also discuss the main modules of a VSR pipeline and the influential datasets. Finally, we introduce some typical VSR application concerns and impediments to real-world scenarios as well as future research directions.
\end{abstract}


\include{content}

\bibliographystyle{unsrt} 
\bibliography{References}

\end{document}

%% file: content.tex
\section{Introduction}\label{sec:introduction}
Naturally, most verbal communication occurs in context when the listener can see the speaker as well as hear them. Although speech perception is normally regarded as a purely auditory process, visual cues can enhance the level of speech understanding~\cite{OuluVS} and McGurk effect demonstrates this influence~\cite{32pdf, mroueh2015deep}. As such, visual information is necessary for deaf people and those who are hard of hearing, to nonverbally and effectively communicate with others.  Probably sign language is an easy and common approach but within the deaf community, lip reading is another method to compensate lack of audio information, to understand speech, and to assimilate with the hearing world.
 
This ability has other numerous  real-world  applications as well, such as biometric identification, visual password, silent speech interface, multi-modal verification systems, forensic video analysis, CCTV\footnote{Closed-Circuit TeleVision} footage~(to assist law enforcement), etc.~\cite{7pdf,LRW, palanivel2008multimodal}. It has also attracted much attention as a complementary signal to increase the accuracy of current Audio-based Speech Recognition (ASR). 

This myriad of potential applications with such capabilities gained the attention toward automatic lip reading systems. In early attempts, image processing techniques have been used, for many years, as feature extractors, but the performance were far from acceptance level for real world scenarios~\cite{Sumby1954}.
Consequently, most of the research efforts in the field of speech understanding focused on ASR systems and underestimated the power of lip reading and visual cues. Nevertheless, in recent years, the principal drivers of innovation in lip reading have been the recent resurgence of deep learning based methods and the great increase in quantity and quality of audio-visual speech datasets, unleashing publication at the scale of tens of impressive works that show the bright future for this research field.

The main objective of this paper is to provide a comprehensive survey of current lip reading methods that benefits from these two ingredients. More specifically, we focus on challenges in automatic lip reading, especially those concerning dataset and feature extraction.  Existing surveys of Visual Speech Recognition(VSR) have only partially reviewed some related topics; For example,  ~\cite{burton2018speaker} have conducted an experiment to determine which lip reading system is  the  most  accurate  for  speaker-independent task.~\cite{LipReadingSurvey} provided a review of traditional and deep learning based architectures grouped by tasks and datasets. 
Particularly, there are three works concentrating on deep learning in VSR: ~\cite{Hao2020ASO}, ~\cite{hao2020survey}, and ~\cite{9522117}.
These works mainly focus on the comparison of various methods and their performance and VSR datasets. This survey, on the other hand, assesses current lip reading methods from a critical point of view to clarify challenges and impediments, those rendering this task more complicated compared to other image and video classification tasks.
More specifically, the outline of the major contributions of this paper relative to the recent literature in the field can be summarized as:
\begin{itemize}
    \item We review the datasets received the highest attention in recent works and their characteristics. We also provide an overview of the chief dataset obstacles presented in the retrospective literature and the corresponding solutions. Moreover, we survey the metrics used for VSR systems evaluation.
    \item For each sub-module of the VSR pipeline, we scrutinize the impediments to progress and to accuracy of the system and then how and to what extent the current methods has removed them or lessened their effects.  
    \item We also present a detailed overview of the open problems and possible future directions.
\end{itemize}

The remainder of this survey is structured as follows.
In Section~\ref{sec:definition}, we define lip reading as a research problem and review usual modules of a VSR pipeline.
In Section~\ref{sec:dataset}, popular datasets, data-related challenges, synthetic data generation methods, and evaluation criteria are summarized.
Recent technical advancements and the progress made in lip reading are also summarized in Section~\ref{sec:automaticlipreading}.
Finally, the possible future directions and the conclusion are presented in Section~\ref{sec:future} and Section~\ref{sec:conclusion}.

\section{Lip Reading: Definition and Pipeline}\label{sec:definition}

\begin{figure*}[h!]
    \centering
    \includegraphics[width=\textwidth]{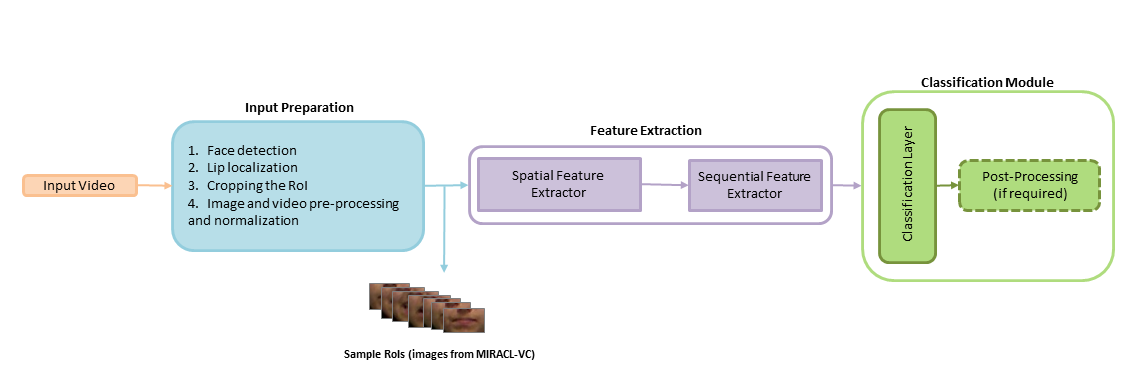}
    \caption{Baseline VSR Pipeline: In a custom lip reading system, the input video usually goes through three sub-modules: Input preparation, Feature Extraction, and Classification. After preparation of the intended RoI, both spacial and temporal features are extracted. And, the final step includes classification and post-processing.}
    \label{fig:vsrpipeline}
\end{figure*}

As a research field, automatic lip reading~\footnote{For simplicity, we use `lip reading' or VSR in the rest of this paper.} can be defined as follows~\cite{Zhou2014, P3D}:

\textit{The process of determining spoken material by analyzing the movements of the speaker's lips in a given video without any acoustic input.}

In other words, lip reading learns a mapping from a video, i.e., sequences of frames/images to a character sequence representing the textual information of a speech. To analyze the input video, it passes through a pipeline of four main steps: input preparation, spatial feature extractor, sequential feature extractor, and classification (Figure \ref{fig:vsrpipeline}). In this section, we briefly introduce them and represent the details in Section ~\ref{sec:automaticlipreading}.

In the first step, the most important task is to select the Region-of-Interest (RoI), which mostly includes the lips and it can be extracted after face detection and lip localization in each frame. The cropped region is then fed to a spatial feature extractor designed to model visual counterpart of characters~(called visemes). This module will focus on changes in lips' shape when uttering different characters. A word, however, is a `sequence' of visemes, thus, the next module is responsible to model the sequential or temporal connection among them.

Spatial and sequential feature extractors capture information required to discriminate among different classes, but to calculate the probability distribution over the output, a classification module is merged into the pipeline where a dense layer with softmax activation is a common option.
According to the level of output, to complete the recognition task, a post-processing method (e.g. a language model) can be considered at this stage as well.

\section {Datasets and Performance Evaluation}\label{sec:dataset}

\subsection{Lip Reading Datasets}
\begin{figure*}[h!]
    \centering
    \includegraphics[width=\textwidth]{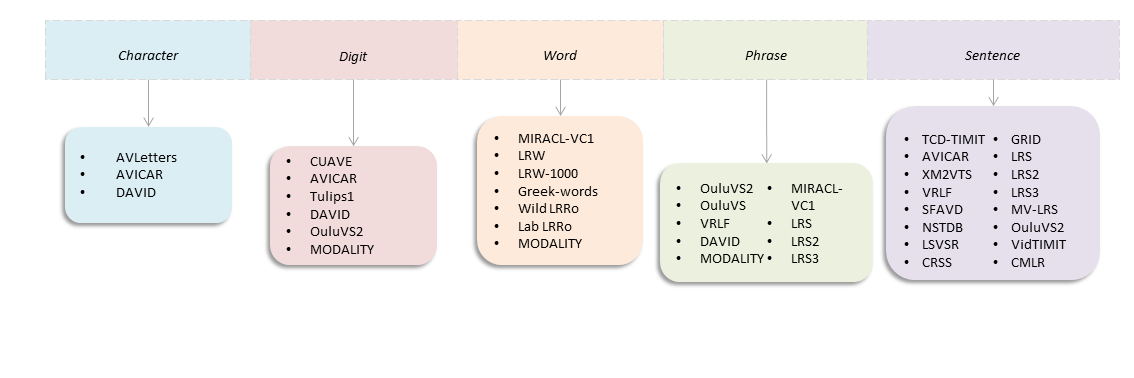}
    \caption{An overview of Lip Reading Datasets. According to the uttered text, datasets can be divided into five categories: character, digit, word, phrase, and sentence. Some of the datasets are included in several categories.}
    \label{fig:datasetsoverview}
\end{figure*}

Based on recording environments/setting, lip reading datasets can be lumped into two categories: (1) controlled setting, and (2) lip reading in the wild. The former includes videos recorded in controlled environments where the position of subjects and the speech content are predefined. The latter attempts to build a dataset from available real world videos, i.e. lectures, debates, etc. Each of these approaches has their own merits. The remainder of this section briefly reviews major datasets collected based on these two settings mostly in English, as the dominant language in current collections, where several well-known ones in other languages are also introduced~(Figure ~\ref{fig:datasetsoverview} and Table ~\ref{tab:table1}). It is worth noting that we can also categorize the lip reading datasets based on the constituents they focus on, i.e., characters, digits, words, phrases, and sentences, as shown in Figure~\ref{fig:datasetsoverview}.  At the end of the section, we also introduce data related issues and possible solutions, the most common methods for generating synthetic samples, and the standard criteria for VSR evaluation.

\subsubsection{Lip Reading in Controlled Environments}\label{sec:controlleddatasets}
Lack of appropriate datasets is probably the preeminent obstacle hindering the advancement of lip reading projects. In other computer vision related applications of deep Convolutional Neural Networks (CNNs), their high performance stems from the large amounts of annotated data. It worth noting the important role of internet as a great source of publicly available information making it possible to build such large-scale collections. Although, the same resource is available for lip reading, there are numerous challenges in annotation and data preparation process leading to inadequacy of visual speech datasets. Probably the first challenge is that the subtitle or text of spoken material in the videos is usually not available except for a small portion of programs. The other challenge is that finding appropriate videos meeting the standard criteria, such as high video resolution, appropriate speaker distance to the camera, considerable speaker variation, to name a few, is not easy. Moreover, the exact start and end of every word or sentence in the video must be clearly marked. These extra efforts of data preparation are probably the biggest impediments to build appropriate datasets for the lip reading task.

An straightforward solution is to record videos in a controlled environment, where a human subject is asked to repeat a set of predefined words or phrases in front of camera. This manner of dataset creation arises from the first application of automatic lip reading: control machines and computers within a specific set of instructions, e.g. voice commands~\cite{DAVID}.
The limited size of vocabulary, clear pronunciations, and recording settings are some important characteristics of such collections~(called controlled datasets). Moreover, due to the limited number of samples, the annotation process would be less laborious for human editors.
In this section, we briefly review well-known datasets recorded in controlled settings.

\paratitle{\textbf{OuluVS.}}
OuluVS\footnote{\url{https://www.oulu.fi/cmvs/node/41315}}~\cite{OuluVS} is an audio-visual phrase-based dataset in which $20$ subjects are asked to sit in a specific distance from the camera~($160$ cm) and repeat $10$ greeting phrases for one to five times. The subjects are from four different countries, accordingly there is a diversity of accents and pronunciation habits. They are also divided in two different groups, where their videos are recorded four days apart. 

\paratitle{\textbf{OuluVS2.}}
OuluVS2\footnote{\url{http://www.ee.oulu.fi/research/imag/OuluVS2/index.html}}~\cite{8pdf} includes the phrase set of OuluVS, together with $10$ randomly generated sequences of $10$ digits, and $10$ randomly chosen TIMIT\footnote{\url{https://catalog.ldc.upenn.edu/LDC93S1}} sentences.
Here, all videos are recorded in an ordinary office environment under mixed lighting conditions and possible background sounds.
The digit sequences are produced once and are the same for all subjects, but the selected TIMIT sentences varies for different trials. The subjects are asked to utter digit sequences and short phrases three times but the TIMIT sentences just once. 
None of the subjects participated in data collection study are native English speakers but they can be grouped into five appearance types: European, Chinese, Indian/Pakistani, Arabian, and African. 

Furthermore, subjects are asked to sit on a chair positioned in a fixed distance of a High Speed (HS) and five High-Definition (HD) cameras located in various positions to record the video from different views. One of the HD cameras and the HS one were exactly in front of the subject~($0^\circ$) and the other four HD cameras were located in different positions: $30^\circ$, $45^\circ$, $60^\circ$ and $90^\circ$ (profile view) to the subject’s right hand side. Thus, models trained on OuluVS2 can potentially be pose invariant.

\paratitle{\textbf{GRID.}}
GRID\footnote{\url{http://spandh.dcs.shef.ac.uk/gridcorpus/}}~\cite{GRID} consists of fixed length synthetic sentences which are generated by the following pattern: (command:$4$; color:$4$; preposition:$4$; letter:$25$; digit:$10$; adverb:$4$) (the numbers indicate the possible options for each part). For instance, `place blue at F $9$ now' is a generated sentence following this pattern. Color, letter, and digit are selected as keywords, and the command, preposition, and adverb are `fillers'. Consequently, there are $1000$ patterns for each subject. The filler words create some variation in contexts for the neighboring key words. Different gross phonetic classes~(nasal, vowel, fricative, plosive, liquid) also were used as the initial or final sounds of the filler words in each position.

GRID was built by participation of $34$ native English subjects including $16$ females and $18$ males. The video and audio are recorded in a single-walled acoustically isolated booth and the camera is located in a static distance from the subject seated in front of a plain blue background.

\paratitle{\textbf{MIRACL-VC.}}
MIRACL-VC\footnote{\url{https://sites.google.com/site/achrafbenhamadou/-datasets/miracl-vc1}}~\cite{MIRACL} is a visual dataset in which $15$ speakers (five men and ten women) positioned in front of a Microsoft Kinect sensor and repeated a set of ten words and ten phrases, each ten times. It contains $3000$ instances, where both 2D images and depth maps are synchronized. The distance between the recording camera and the speaker is about one meter. MIRACL-VC can be used for a variety of research fields like face detection and biometrics. 

\paratitle{\textbf{VidTIMIT.}} 
As the name suggests, the VidTIMIT\footnote{\url{http://conradsanderson.id.au/vidtimit/\#examples}}~\cite{33pdf} is also based on TIMIT sentences. It includes video and audio recordings of $24$ males and $19$ females. Each subject recited $10$ TIMIT sentences chosen from the test section of the TIMIT corpus. The data was recorded in three different sessions with delays of seven days between first and second sessions, and six days between second and third sessions. The first two sentences for all subjects are the same, with the remaining eight generally different ones for each subject.
The recording was done in an office environment using a broadcast quality digital video camera. 
Besides automatic lip reading, VidTIMIT can be useful for research on topics such as multi-view face recognition, multi-modal speech recognition, and person identification.

\paratitle{\textbf{TCD-TIMIT.}} 
TIMIT sentences are also used for building another audio-visual dataset named. TCD-TIMIT\footnote{\url{http://www.mee.tcd.ie/~sigmedia/Resources/TCD-TIMIT}}~\cite{TCD}. The speakers in this dataset can be divided in two groups of so-called volunteers (normal-speaking adults) and the professionally trained lip speakers who attempt to make their mouth movements more distinctive and to provide insight as the best features to use for visual speech recognition.
The video modality is provided in two angles of $0^\circ$ and $30^\circ$ (the camera is positioned on the speaker’s right side). The frontal view offers information about mouth width, but the profile one offers information about lip protrusion. 
In addition to the text annotation corresponding to the spoken sentences, a phoneme-to-viseme mapping is provided.

\paratitle{\textbf{CUAVE.}}
In Clemson University Audio-Visual Experiments (CUAVE)  corpus~\cite{CUAVE, CUAVE_2}, the participants recited connected (zero through nine) and continuous digit strings (like telephone numbers) in an isolated sound booth, where $36$ subjects ($17$ females and $19$ males) repeat $50$ connected digits while standing still in front of a camera~(resulting in $50\times36$ utterances). Moreover, they also intentionally moved side-to-side, back and forth, and tilted their head while speaking $30$ connected digits (resulting in $30\times36$ utterances). The same configuration was also applied for continuous strings, i.e., the speakers uttered three phone numbers while sitting stationary ($30\times36$ utterances) and three others when moving ($30\times36$ utterances). The video of both profile views (left and right) are recorded when the subjects are repeating $10$ connected digits(resulting in $2\times10\times36$ utterances).

CUAVE also provides videos of $20$ pairs of speakers, which is helpful in multi-speakers setting. These videos are valuable in experiments of distinguishing the speaker from others and to recognize speech from two talkers. The latter is challenging based on audio information only. Hence, the video modality will assist the recognition task. In this setting, the first speaker repeats the continuous-digit sequence, followed by second speaker and vice versa (two sequence for each speaker) and the third sequence is uttered when both speakers talk simultaneously.

\paratitle{\textbf{AVICAR.}}
As its name suggests, the Audio-Visual Speech Recognition in a Car (AVICAR) dataset\footnote{\url{http://www.ifp.illinois.edu/speech/sst/AVICAR}}~\cite{AVICAR} has been recorded inside a car, where four cameras are placed on the dashboard. AVICAR consists of audio and video files as well as the text annotation of isolated digits, isolated letters, ten-digit phone numbers, and TIMIT sentences uttered by $100$ subjects. This dataset can be served for the purpose of automatic dialing and the study of homophones. Moreover, phonetically balanced sentences randomly selected from $450$ phonetically compact sentences of the TIMIT speech database, are included to provide training and test data for phoneme-based recognizers.
In AVICAR, $60\%$ of subjects are native American and the others have Latin American, European, East Asian, and South Asian backgrounds. The speakers are divided in $10$ groups of five males and five females. For each group, a different script set is prepared, where $118$ utterance are recorded for each script set.

\paratitle{\textbf{CRSS-4ENGLISH-14.}}
CRSS-4ENGLISH-14~\cite{tao2018gating} comprises utterances with various length, from single words (e.g., “Worklist”), and cities (e.g., “Dallas, Texas”) to short phrases or commands (e.g., “change probe”), continuous numbers (e.g., “4, 3, 1, 8”), continuous sentences (e.g., “I’d like to see an action movie tonight, any recommendation?”), and questions (e.g., “How tall is the Mount Everest”). CRSS-4ENGLISH-14 is a controlled dataset collected by the Center of Robust Speech System (CRSS) at The University of Texas at Dallas (UTDallas).
The subjects were asked to utter the requested token in a 13ft ×13ft American Speech-Language-
Hearing Association (ASHA) certified sound booth, illuminated by two professional LED light panels and equipped with multiple microphones and cameras. The participants were also asked to repeat randomly selected tasks when a prerecorded noise (of mall, home, office, or restaurant) is played. The final recording length for each subject is 30 minutes and is transcribed manually. 
The speakers accents fall into 4 categories: American (115), Australian (103), Indian (112) and Hispanic (112) and average age among subjects is 25.58.

\paratitle{\textbf{Small-Scale datasets.}}
Thus far, we have reviewed large-scale datasets, however, there exist several small datasets useful for model evaluation and specific applications. We now briefly discuss them in following. 

- \emph{DAVID}: This dataset includes the audio and video recordings of English alphabets, isolated digits, application-specific control commands, and nonsense utterances of consonant and vowel sounds sequences~\cite{DAVID}. 

- \emph{MODALITY}\footnote{\url{http://www.modality-corpus.org/}}: It provides high-resolution RGB and depth images recorded in an acoustically adapted room. The videos are recorded by two cameras placed at $30$ cm and $70$ cm of the speaker and a depth camera located at $40$ cm of them~\cite{27pdf}. Almost $50\%$ of the speakers are English natives. The language material is selected to reflect the frequentation of speech sounds in Standard Southern British so that it can be useful in vowel recognition studies. The utterances cover the numbers, names of months and days, and a set of verbs and nouns mostly related to controlling computer devices.

- \emph{Tulips1}: It is the first dataset recorded for the purpose of lipreading and includes the videos of the first four English digits, where $12$ subjects repeat each of them twice~\cite{Tulips1}.

- \emph{AVLetters}: As an audio-visual database, AVLetters~\cite{AVLetter} consists of $780$ utterances of isolated letters A-Z by $10$ speakers (five males (two with moustaches) and five females). Each speaker repeats the letters for three times.

- \emph{XM2VTSDB}\footnote{\url{http://www.ee.surrey.ac.uk/CVSSP/xm2vtsdb/}}: It includes the frontal face recordings of subjects uttering three specific sentences. The data is recorded in four different sessions, when every sentence is repeated twice~\cite{XM2VTSDB}.

\subsubsection{Lip Reading Datasets in the Wild}\label{sec:wilddatasets}

The benefits of datasets recorded in controlled environments notwithstanding, they have several characteristics restricting their applications and benefits for research in VSR field, such as limited number of samples in each class and lack of diversity in subjects~\footnote{There might be considerable number of samples in each class, but every subject usually repeats each spoken unit for several times.}. 
As a result, the trained models have high performance on development and test sets of the same collection but fail when processing videos recorded in real world conditions due to differences in illumination, the speaker pose, and pronunciation distinctiveness. Additionally, it is mostly impossible to precisely annotate the boundaries of the spoken unit in other videos in test time and consequently the inference accuracy drops drastically. The videos recorded in controlled environments can be used as a benchmark for evaluating VSR pipelines, to pre-train models, and speed up the convergence of the training procedure, but the need for creating proper lip reading data from videos recorded in a non-supervised scenario or generally speaking in the wild, is inevitable. Thus, an appropriate dataset preparation technique must be designed to automate the annotation process.  

\begin{figure*}[h]
    \centering
    \includegraphics[width=\textwidth]{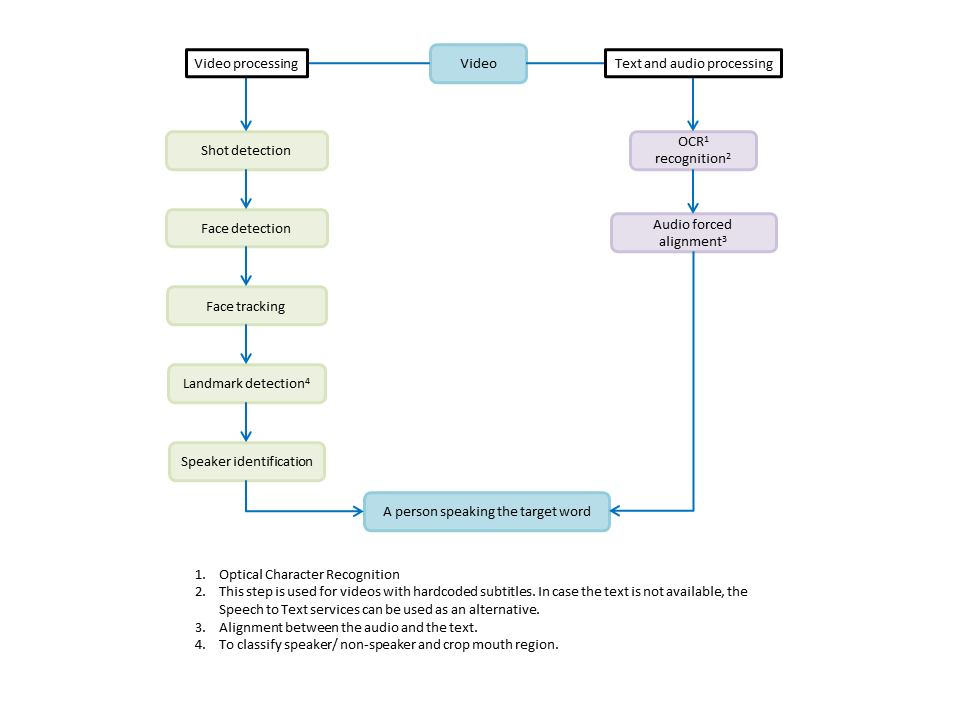}
    \caption{Multi-stage pipeline for automatically collecting and processing a large-scale audio-visual dataset}
    \label{fig:datasetpipeline}
\end{figure*}

For the first time, Chung and Zisserman~\cite{LRW} proposed such a pipeline to produce samples from BBC\footnote{British Broadcasting Corporation} news videos. Consequently, other researchers adopted a similar approach on videos of various resources such as TV series and Youtube videos to build lip reading datasets. The main steps of this pipeline is summarized in figure \ref{fig:datasetpipeline}, which is explanatory enough so that we skip the details.

The automation of video annotation has reduced the human editors effort to a great degree and has resulted in several datasets that not only have made the trained models more accurate but also robust and reliable. In following, we review some of the well-known English datasets in the wild.

\textbf{Lip Reading in the Wild (LRW)}

The LRW\footnote{\url{http://www.robots.ox.ac.uk/~vgg/data/lip_reading/lrw1.html}} dataset~\cite{LRW} is one of the established English word-based audio-visual lip reading datasets collected by the Visual Geometry Group(VGG) at Oxford university. This dataset is produced based on the multi-stage pipeline (Figure \ref{fig:datasetpipeline}) and covers BBC productions including the news and debate programs. The reason for this selection is that they cover a wide range of speakers, the shot changes are less frequent, and the speakers usually talk without interruption. Therefore, there are more full sentences with continuous face tracks. 

The data is divided into three sets of train, test, and validation that are disjoint in time. Table~\ref{tab:table1} lists the statistics and characteristics of LRW dataset. It is the largest word-based dataset, and has three characteristics rendering it challenging.

First, each instance does not completely represents a single isolated word and there may be co-articulation of the lips from preceding and subsequent words. It may appear counterintuitive at first that this weakly annotated word boundaries can help to increase the accuracy and robustness of the model, but the experience of using LRW for training showed that the models have learned not only to discriminate the $500$ target words but also to spot each of them in the video and ignore the irrelevant information~\cite{Important}. As emphasized by~\cite{Important}, this feature might be helpful only in sentence level analysis, however, it can also be advantageous for real-world applications, when it would be difficult to spot the words. Some random examples of utterances are `...the (election) victory...', `...and so (senior) labour...' and `...point, I (think) the ...' (the words in the parentheses are the labels). 

Second, the length of samples are $29$ frames and the target word is always uttered at the middle of the video~\cite{18pdf}. It is worth noting that the trained model on this biased setting becomes sensitive to tiny changes being applied to the input. For example, by removing some random frames or shifting them, the model fails at spotting and classifying the video since it has learned to look for the occurrence of the target word in the middle of the sample. To address this problem, variable length augmentation can be used as a solution~\cite{18pdf}, where the input sequence of training set is cropped temporally at a random point prior and after the target word boundaries.

Third, target words in LRW do not include short words due to homophones. These types of words share the same visemes but are actually different words such as `bad' and `bat'~\cite{homophenes}. Homophones can make the lip reading process an arduous task not only for VSR systems but also for human. Although short length homophone are avoided, there are $23$ pairs of singular and plural forms of the same words (`benefit' and `benefits'), and $4$ pair of present and past forms of some regular verbs (`allow' and `allowed')~\cite{11pdf}.

\textbf{Lip Reading Sentence (LRS)}

A similar processing pipeline used in LRW is also applied for creating LRS~\cite{LRWS} as an audio-visual dataset by VGG. The videos are divided into sentences and phrases based on the punctuation marks in the transcripts. Each sentence contains $100$ character and the corresponding video lasts for $10$ seconds. It includes a variety of BBC programs recorded between $2010$ and $2016$ and the selection is based on the aforementioned reasons for LRW. The provided content is also divided into non-overlapping train, validation, and test sets according to broadcast date.

\textbf{LRS2-BBC}

Another audio-visual speech recognition dataset based on BBC programs is LRS2-BBC\footnote{\url{http://www.robots.ox.ac.uk/~vgg/data/lip_reading/lrs2.html}}~\cite{12pdf} which provides utterances in both sentence and phrase level. The process of program selection for LRS2-BBC includes various type of programs, unlike LRW and LRS which only focus on news and debate programs. Consequently, LRS2-BBC has more sentences and phrases than LRS. The same data production pipeline and video/character length constraints of LRS are applied to LRS2-BBC as well.  

The LRS2-BBC has a pre-train set along with development (train/val) and test sets divided according to broadcast date. The pre-train set contains sentence excerpts which may be shorter or longer than the full sentences included in the development set, and are also annotated with the alignment boundaries of every word. 

\textbf{LRS3-TED}

VGG group collected a dataset based on the videos of TED and TEDx talks in English. 
LRS3-TED~\footnote{\url{http://www.robots.ox.ac.uk/~vgg/data/lip_reading/lrs3.html}}~\cite{LRS3} meets the program selection criteria stated for LRW and LRS. The videos are clipped to $100$ characters or six seconds. One important characteristic of LRS3-TED is that subjects in test, train, and validation sets are mostly unidentical. The identities are not labeled manually but it is unlikely that the same speaker appears on TED programs repeatedly.
On the other hand in LRW, LRS, and LRS2-BBC that cover regular TV programs, the same speakers are likely to appear in common from one episode to the next, so that test, train, and validation sets may have some identities in common and consequently they are not speaker isolated.

\textbf{Multi-View Lip Reading Sentences (MV-LRS)}

MV-LRS~\cite{17pdf} is an audio-visual speech recognition dataset, based on LRS, collected by VGG with two main differences. First, it includes faces from different views, frontal to profile, and second, in contrast to LRS which mostly covers broadcast news, MV-LRS includes a wider range of resources such as dramas and factual programs, where speakers engage in conversations with one another and are therefore more likely to be pictured from various views. The data preparation pipeline is as same as the one used in~\cite{LRW}. The videos are also divided into train, validation and test sets according to dates used to split LRS.

\textbf{Large-Scale Visual Speech Recognition(LSVSR)}

LSVSR~\cite{1pdf} is an order of magnitude greater than prior datasets built from Youtube videos. In addition, the content is much more varied (i.e. not news-specific), resulting in a larger vocabulary size. The data preparation pipeline is similar to Figure~\ref{fig:datasetpipeline}, however, it uses face matching techniques for face tracking instead of traditional methods (KLT tracker~\footnote{Kanade–Lucas–Tomasi feature tracker}) used in LRW. 

\subsubsection{Non-English datasets}
Although many lip reading datasets are in English, there are few efforts to build collections for other languages, as listed in Table \ref{tab:table1}. For example, LRW-1000\footnote{\url{http://vipl.ict.ac.cn/en/view_database.php?id=13}}~\cite{D3D} is a word-level Mandarin dataset. It has $1000$ classes labeled with English letters; each of them corresponds to a syllable of a Mandarin word composed of one or several Chinese characters. 

News, Speech, Talk show DataBase (NSTDB)~\cite{26pdf} is another Mandarin corpus created based on CCTV\footnote{China Central Television} News and Logic Show. Since many of Chinese characters have the same pronunciation, Hanyu Pinyin~\footnote{Hanyu Pinyin is the Mandarin romanization system.} without tone is selected as the label of each class, leading to $349$ categories in this sentence-level dataset.
Another Mandarin dataset created based on Chinese news program is Chinese Mandarin Lip Reading (CMLR)~\cite{CMLR}. This sentence-level dataset comprises over 100,000 natural sentences, which are extracted from China Network Television website recorded between June 2009 and June 2018. 

Wild LRRo and Lab LRRo are two Romanian lip reading word-level datasets~\cite{LRRo}. The former is generated from videos of Romanian TV shows, TV news programs and Romanian TEDx talks. However, the later includes simultaneous recordings of frontal and $30^\circ$ (left side) views in a controlled environment.

Sharif Farsi Audio Visual Database (SFAVD)~\cite{SFAVD}, Greek words\footnote{\url{https://github.com/dimkastan/LipReadingGreekWords}}~\cite{Greek}, and Visual Lip Reading Feasibility (VLRF)\footnote{\url{http://fsukno.atspace.eu/Data.htm\#VLRF}}~\cite{VRLF}, are also three non-English datasets in Farsi, Greek, and Spanish, respectively. 

\subsection{Data Challenges}\label{sec:difficulties}

Lip reading is a challenging task by its nature; the differences between consecutive visemes are trivial and the uttered text must be inferred from the non-obvious motions in the given sequence. On the other hand, there are variety of factors making test videos different from development data, and consequently overshadowing the accuracy and performance of lip reading systems.
In this section we review some of these data related issues, their source, and the common solutions to mitigate their effects.

The factors fall into three broad categories: (1) subject-dependent, (2) video quality, (3) and content-based. 

\paratitle{\textbf{Subject-dependent factors}}:  
This category includes the factors increasing the intra-class variance. Pronunciation, accents, speaking rates, facial hairs, and shape of lips are some important individual features falling into this category. Furthermore, people may mumble, have different facial expressions while speaking, or move their heads to express their feelings. 
These speaker-related features and behaviours can only be addressed by increasing the speaker's diversity in the dataset.

\paratitle{\textbf{Video quality factors}}:
Video quality factors are quite utterance independent. In particular, these factors  are basically related to the recording settings and can be subdivided into smaller groups of: 
\begin{itemize}
    \item Camera's specification: frame-rate and resolution.
    \item How the speaker is exposed to the camera: the subject's distance to the camera, head orientation or the camera's angle toward speaker, occlusion distortions (having mask over face or covering the mouth), and subject's head movement causing motion blur.
    \item Visual distortions: digitization artifacts, noise corruption, filtering, lighting condition, etc.
\end{itemize}

The most common solutions to mitigate the effects of these factors are temporal sampling, image normalization and resizing, data augmentation, multi-view samples, and videos recorded in various environmental conditions.

Artificial perturbations~\cite{su2019one} is a common attack to fool a DNN and can also be classified as video quality related factors. To increase the model's robustness against such attacks, data augmentation is a straightforward solution. Stability training approach~\cite{zheng2016improving} can also help the model to learn more robust feature embeddings. 

\paratitle{\textbf{Content-based factors}}:
These factors are associated to the quantity and quality of the samples from lip reading point of view. Inter-class similarity (same viseme but different phoneme (`b' vs. `p') or different words but the same lip movements (`bat' vs. `bad'))~\cite{39pdf}, lack of enough utterances in each class, and not so clean data (e.g. videos also contain other words along with the one offered as the label) are among the most important content-based factors. Probably, increasing the number of samples in each category can attenuate some of these issues.

\subsection{Synthetic Datasets: Solution for Data Concerns}\label{sec:generation}

We already pointed out that lack of enough, well-annotated data is the main challenge in developing VSR systems. For example, LRW is the only word-based collection in English. This challenge is probably the case for other image and video related tasks as well. As a solution, in recent years, Generative Adversarial Networks (GANs)~\cite{Goodfellow2014} have proven to be effective to synthesize realistically looking images that can be employed for research and benchmarking purposes~\cite{Osokin2017,5pdf}.
In the field of speech-driven facial animation,~\cite{Vougioukas2018} deployed Temporal Conditional Generative Adversarial Networks (TC-GANs) to produce realistic talking heads. In this method, the generator accepts a still face image and an audio signal which is sub-sampled into small overlapping chunks.~\cite{5pdf} employed similar approach to construct lip reading data from a still image of lips and an audio of the text unit that is supposed to be spoken\footnote{If the audio sample file is not available in the dataset (unseen class), an off-the-shelf Text-To-Speech software (TTS) is used to generate one.}. In generator, the image and audio modalities are fed to separate encoders `Identity Encoder' and `Context Encoder', respectively. Next, the concatenation of these encodings along with a noise component generated by the `Noise Generator', builds the latent representation which is passed to the`Frame Decoder' to produce  the  video  frames.  
Finally, the quality of generated video frames is evaluated by two different discriminators: the `Frame Discriminator' to assess the quality of generated lips and the `Sequence Discriminator' to ensure that the generated video exhibits natural movements and is synchronized with the audio.

This data generation technique is useful to increase the samples of each existing class as well as producing utterances for new unseen classes. However, visemes concatenation is another solution for the later task~\cite{Huang2002}.
In this technique which is employed to generate photo-realistic talking head sequences of unspoken words, the text of utterance is fed to a Text-to-Speech module and then CMU\footnote{Carnegie Mellon University} Pronouncing Dictionary~\cite{phonemizer} to generate a sequence of phonemes. Next, the visual counterpart of each phoneme is extracted using Jeffers phoneme-to-viseme map~\cite{JeffersJanet1971} and speaker-to-viseme database. Finally, the concatenation of the resulting visemes forms the video. Note that the speaker-to-viseme database is manually annotated  for each subject. 
Moreover, this method is useful for phonetically close languages. For instance, ~\cite{5pdf} used the phonemes similarity between English and Hindi to generate lip reading data in Hindi.

To compare the quality of samples generated using the mentioned approaches,~\cite{5pdf} employed both of these techniques to generate synthetic videos for unseen classes. The models performance showed that viseme concatenation is not as effective as GAN-based method. This result is probably due to the fact that for GAN-produced videos, both frame and sequence discriminators ensure the quality and naturalness of the output. On the other hand, in the viseme concatenation technique, although the frames represent the natural lip's shape when pronouncing specific character, there is no module to check the smoothness of the generated viseme sequences.

Except for generating samples in both seen and unseen classes, creating synthetic data is also used to answer other data concerns in the field of lip reading.
As we stated in previous section, in natural conversations, speakers head movement during speaking is another data concern affecting the performance of VSR systems.
With what we have presented so far, most of the datasets are collected based on the hypothesis that the speaker has the least head movement; the fact that contradicts with real conversations that do not take place in studios like datasets collected in controlled environments. Consequently, the accuracy of the model trained on frontal or near frontal faces drops drastically in real-world applications~\cite{pose_invarient}.

To mitigate this problem, a solution used in controlled datasets is to record data from different views similar to OuluVs2. The models trained on such collections outperform those trained on the single-view videos, and the best results achieved when transferring knowledge of other views data in to a unified single view.
Despite this improvement, the variation of head poses in such datasets is limited and thus the accuracy flaw still remains. For example,  MV-LRS, as a dataset in the wild, can cover the deficiencies of OuluVs2, head pose still remains a challenge.

An approach is to generate synthetic non-frontal data from frontal or near frontal view videos of existing visual speech collections. To do so, 3D Morphable Model (3DMM)~\cite{Egger2019} as a data augmentation technique is an appropriate solution to generate synthetic samples in arbitrary poses.
3DMM is a type of generative model that creates samples by applying randomized shape and albedo deformations to a reference mesh~\cite{Sutherland2020}. Thus, 3DMM can be an effective way to produce videos covering a full and continuous range of poses.
Using this technique,~\cite{pose_invarient} proposed a pose augmentation pipeline to generate word-level lip reading data. This method fits the 3DMM into each frame from the video, estimates the 3D pose of detected face, and then randomly select two pose increment angles, one in yaw ($-45^\circ$ to $45^\circ$) and another in pitch ($-30^\circ$ to $30^\circ$) direction. This shift is applied on all video frames and then results are rendered to produce the new video.
The experiment showed that incorporating this method in the training pipeline along with 2D image augmentation (random noise, random crop, etc.) can improve the final word accuracy.

\subsection{Evaluation Criteria}

Various metrics have been utilized to evaluate the performance of VSR systems, including word accuracy~\cite{LRW} and Sentence Accuracy Rate (SAR)~\cite{37pdf}. 

Error Rate (ER) metrics on different levels such as word, character or viseme are another family of evaluation criteria in the field of lip reading. In these metrics, the decoded text is compared to the actual text and the overall distance is computed, as stated in following,

\begin{equation}
   ER  = \frac{S+D+I}{N}
\end{equation}

where $S$, $D$ and $I$ are the number of substitutions, deletions, and insertions, respectively, to convert the predicted text to the ground-truth, and $N$ is the total number of tokens (i.e., characters, words, visemes) in the ground-truth. $ER$ changes as a function of test sentence length~\cite{12pdf} which is an important characteristic when comparing different approaches.

BiLingual Evaluation Understudy (BLEU) score~\cite{LRWS} is another metric used to evaluate the performance of lip reading systems. BLEU, which originally is developed for machine translation performance evaluation, is a modified form of $n$-gram precision to compare a candidate sentence to one or more reference sentences~\cite{BLEU}, and is calculated as follows:

\begin{align}
\nonumber
    BLEU & = \frac{CountClip}{CN} \\
\end{align}

\begin{align}
\nonumber
    CountClip & = \min (Count, MRC)
\end{align}

where $CN$ is the number of $n$-grams in candidate text sequence, $Count$ is the intended word’s count in the sentence, and $MRC$ (Maximum Reference Count) denotes the maximum number of intended word repetition in all the references. A perfect match have BLEU score $1.0$, whereas a perfect mismatch score is $0.0$.

\section{Automatic Lip Reading}\label{sec:automaticlipreading}

\subsection{Input Preparation}

In this module, face detection and lip extraction are the very first tasks.  
Before popularity of deep learning based landmark prediction models, traditional approaches often used color information or structural information for lip detection~\cite{hao2020survey}, but pre-trained deep models, such as Dlib~\cite{king2009dlib} and RetinaFace~\cite{deng2019retinaface}, have made this process faster, more accurate, and easier to integrate in any VSR pipeline. The lips region is generally selected as the input to the VSR system, however, several studies have demonstrated that their changes are not the only visual signal helping to decode speech~\cite{Sumby1954}. For instance,  movements of and changes in extra-oral facial regions, such as the tongue, teeth, cheeks, and nasolabial folds~\cite{RoI} during speaking, can also assist VSR specially when normal sound is not available. A comparative study on the RoI selection attests that including the upper face and cheeks increases the VSR's accuracy constantly~\cite{RoI}.

Pre-processing and normalization are the next steps in input preparation module. The video samples of different classes usually have different length, as well as various resolutions and different frame rates are inevitable. Thus, to feed inputs with same shapes to the network, image resizing and frame sampling are necessary. Moreover, image normalization is an important approach to make the convergence faster while training the network.

\subsection{Feature Extraction}
As we stated in Section \ref{sec:definition}, after input preparation, the RoI is fed to the feature extraction module. In this section, we illustrate these feature extractors and how they have evolved to achieve better lip reading performance.  

\subsubsection{Spatial Feature Extractor}

The main focus of this survey is lip reading systems in deep learning era, accordingly in this section, we first introduce conventional approaches (e.g., handcrafted features) and then mainly focus on deep learning architectures for spatial feature extractors.

\paragraph{Classical Methods}

To extract  handcrafted features, some well-known approaches are image transform techniques, motion features, mouth geometry, statistical models, and texture-based methods. For example, many researchers have proposed VSR systems based on `image transform techniques' such as Principal Component Analysis (PCA) and Discrete Cosine Transform (DCT)~\cite{lucey2008patch, Stewart2008, OuluVS, Gowdy2004, Saenko, 30pdf, lee2008robust, lucey2008patch, Ivana2006, Basu1999, tao2018gating}. 
Texture-based methods, such as Local Binary Pattern (LBP), are also popular in lip reading systems to capture the local changes of frames in spatial and temporal domains~\cite{OuluVS, Zhou2011}.
Shape Difference Feature (SDF)~\cite{wu2016novel} is also another visual feature based on the geometry of lip in each viseme, extracted based on lip width, height, and contour.

We refer the readers to~\cite{LipReadingSurvey} and ~\cite{hao2020survey} for more information on classical approaches. 

\paragraph{Deep Learning Based Feature Extractors}\label{sec:visualfeatureextractor}

\begin{figure*}[h]
    \centering
    \includegraphics[width=\textwidth]{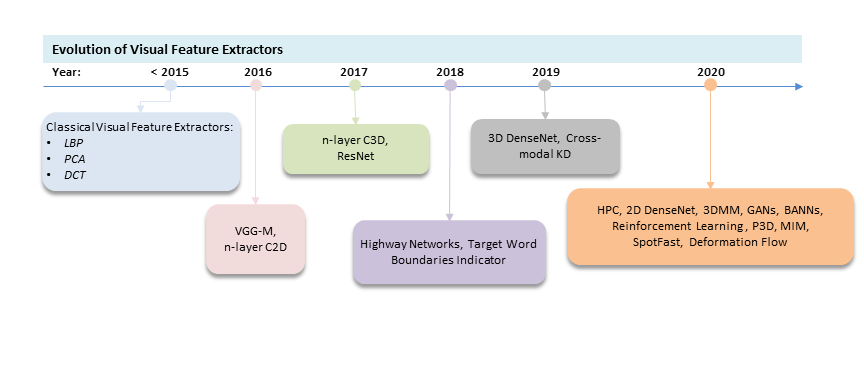}
    \caption{Evolution of Spatial Feature Extractors in Lip Reading Pipeline from 2015 to 2020.}
    \label{fig:visualfeatureExctractor}
\end{figure*}

In lip reading context, the changes happening in consecutive frames are subtle (specifically for words with less mouth movements) and the performance of VSR systems based on hand crafted features proves that they fail to model these details and the discrimination among classes.
A deep-based alternative method is a dense layer to render the high dimensional input image into a low dimensional representation~\cite{Oulu}. However, the results were not promising as they often reduce the dimensionality and probably miss the important details.
On the other hand, the emergence of CNNs~\cite{alexnet} and their success in various computer vision tasks demonstrate that with the help of a large-scale dataset, they can learn robust and powerful feature representations~\cite{object_detection_survey}. 
Depending on their depth and breadth, CNNs are capable of extracting features at different levels, from high-level ones that can represents semantic meanings, to low-level local features that are common among samples ~\cite{obj_detection_survey}. 
Now, with the help of large-scale visual speech collections, 2D and 3D CNNs seem to be feasible solutions to enhance the capacity of models.
In this section, we mainly focus on the lip reading systems based on these networks and review their features from various points of views, including what types of features they extract (spatial or spatio-temporal), the key techniques, and novelties. A review of methods covered in this paper is available in Table \ref{tab:table2}. Additionally, in Figure \ref{fig:visualfeatureExctractor}, the evolution of spacial feature extractors after popularity of deep-based methods is illustrated. 

\paratitle{\textbf{2D Feature Extractors.}}

In computer vision, 2D CNN (C2D) is generally a good choice to deal with spatial information. But, its application in lip reading does not follow a unified philosophy. One common approach is to apply 2D convolutions on each frame and extract lip's discriminative features~\cite{LCANET, D3D, CMLR, zhao2020hearing, petridis2018audio, tao2020end}. This method, known as Multiple Towers (MT), pays attention to the shape of lips when uttering a specific character. 
An alternative approach is Early Fusion (EF) in which the frames are stacked and then fed to a C2D. The early and direct connectivity to pixel data allows the network to precisely detect local motion direction and speed~\cite{karpathy2014large}. EF approach tries to capture local temporal or motion features, as well as spatial ones. 

Chung and Zisserman~\cite{LRW, CHUNG201876} used EF and MT in their proposed word-based VSR methods. 
In EF, a 2D network, similar to the first three blocks of VGG-M, ingests a \textit{T}-channel image, where each of the channels encodes an individual gray-scale frame of the input video. VGG-M~\cite{VGGM} is selected due to its good image classification performance, faster training procedure, and  memory-efficient characteristic.
On the other hand, in MT, each frame goes to one of the \textit{T} towers with similar architectures to the first block of VGG-M (with shared weights among towers). The features are then concatenated channel-wise, a $1\times1$ convolution is then applied to reduce the number of channels, and finally the extracted features are fed to the second and third blocks of VGG-M. 
The results show that using 2D convolutions in MT fusion mode is more effective than EF and provides superior performance compared to the handcrafted features by a significant margin.
The success of MT also emphasizes the fact that it is more effective to capture motion among high level features instead of frames.

Watch, Listen, Attend and Spell (WLAS)~\cite{LRWS} and Multi-view Watch, Attend and Spell (MV-WAS)~\cite{17pdf} also adopt MT strategy in their spatial feature extractors. In the proposed models, the video input is fed to the \textbf{Watch} module; a VGG-M network ingesting a sequence of five frames in gray scale and reverse order. MV-WAS follows the same strategy to process the video modality. However, it takes a larger RoI (the whole face bounding box) with higher image resolution.

\paratitle{\textbf{3D Feature Extractors.}}

In contrast to C2D, 3D CNN (C3D) is able to process the input in both spatial and temporal domains, simultaneously. Similar to EF, this type of feature extractors also capture short temporal features corresponding to frames representing the pronunciation of the same character. In contrast to applying 2D convolution on an image and a video volume (multiple frames as multiple channels) which results in an image, applying 3D convolution on a video volume results in another volume, preserving temporal information of the input signal~\cite{Tran2015}.

Basic 3D feature extractors~\footnote{In the rest of this paper, vanilla C3D or simply 3D convolutions refers to a stack of basic 3D convolutions.} are mainly used for sentence-level prediction. For example, LipNet~\cite{LipNet} employs 3D convolutions and pooling layers with different kernel sizes to extract features at different levels. Instead of stacking simple 3D convolutional blocks, another popular and successful approach among VSR pipelines is to deploy 3D counterparts of well-known 2D networks. Vision to Phoneme (V2P) network~\cite{1pdf} is a good example that employs 3D version of VGG, outperforming LipNet and visual modality of WLAS on LSVSR. 
~\cite{D3D} also trained 3D DenseNet (D3D) for word-level prediction. DensNet~\cite{DensNet} is a well-known 2D residual network introduced to solve the problem of vanishing gradient and over-fitting in image classification. It establishes a connection between different layers and utilizes shallow features with low complexity. In Densenet and similar residual networks, passing information from one layer to the next makes it possible to substantially increase the depth of the network and achieve more training efficiency and accuracy~\cite{He2015}. 

Instead of residual networks, another approach to overcome vanishing gradient is Highway Networks~\cite{highway}. Inspired by Long Short Term Memory units (LSTMs) and as a pioneer technique to effectively train end-to-end deep networks, Highway Networks employ a learned gating mechanism for regulating information flow. 
A recent study on language modeling~\cite{Kim2015} shows that, by features combination, Highway Networks increase the modeling capability of the encoder and result in richer semantic features. This emphasize the potential of investigating the effectiveness Highway Networks for lip reading task.

LCANet~\cite{LCANET} is a VSR model benefiting from this type of networks. With this design, the model has paths along which information passes through several layers without attenuation. Highway Networks along with 3D CNN and Bidirectional Gated Recurrent Units (Bi-GRU) are parts of a spatio-temporal video encoder that captures long-term and short-term features and encode more complex semantic information.
 
Despite the fact that deep models result in more accurate lip reading, they are not memory and time efficient in both development and test phases and cause impediment for real-time applications. A modification that can be applied on C3D to alleviate these problems is depthwise separable convolution~\cite{Mobilenet} in which a $N \times N \times N$ convolution is separated into $1 \times N \times N$ convolutional filters on spatial domain and $N \times 1 \times 1$ convolutional filters on temporal domain.
Using such convolutions, the model is computationally more effective and has less parameters compared to the vanilla C3D counterpart (less Floating Point Operations (FLOPs)).
Audio Enhanced Multi-modality Speech Recognition network (AE-MSR)~\cite{P3D} is a good example using depthwise separable convolutions in which the visual features are extracted by a Pseudo-3D Residual Convolution (P3D) sub-network. The proposed light model is a $50$-layer P3D network that cyclically integrates the three versions of P3D blocks. Each version employs different strategies to apply convolutions and residual connections on the input tensor. The final visual feature extractor is composed of a C3D block, batch normalization, ReLU activation, and max-pooling layers followed by a $50$-layer P3D ResNet.

\subsubsection{Machine Learning Strategies and Novelties}

After reaching a plateau in classification accuracy, the network architecture design for spacial feature extractors was quickly gravitated towards strategies and novelties in applying convolutional kernels, heuristics, modalities, etc. Therefore, this section is dedicated to these retrospective research efforts.

\paratitle{\textbf{Lip Reading as a Video Classification Task.}}

Considering lip reading as a video classification or more specifically action classification task is a common approach~\cite{LRW}. Nevertheless, it is worthwhile noting that such architecture designs have low chance of being effective for lip reading due to its specific characteristics and  challenges. For instance, in video-based action classification, there might be coarse-grained features such as a huge shift in the number and types of the objects in the frames, their positions according to each other, background, etc. Accordingly, global pooling which is usually regarded as a structure regularizer is beneficial and can explicitly enforces feature maps to be confidence maps of categories. However, in a video representing lip regions, there is no such changes in the scene to make them differential.
Furthermore, in lip reading task, 3D global pooling is commonly employed for 3D feature extractors, applied to both spatial and temporal dimensions~\cite{1712.01111}. As demonstrated by~\cite{Tran2015}, the temporal pooling is important for recognition task to better model the spatio-temporal information of video and reduce background noises. Although, when the difference among visemes are subtle but crucial applying global pooling might results in loss of information in both spatial and temporal domains~\cite{24pdf}. Moreover, this technique consumes local spatial information critical to capture the subtle changes in the appearance and the state of the lips, thus, activation in different spatial locations, which correspond to different visemes, may contribute the same to the final features. 

To overcome the mentioned challenges,~\cite{24pdf} substitute Spatio-Temporal Fusion Module (STFM) for global average pooling and dimensionality reduction. In STFM, a spatial pooling operation similar to RoI-Pooling~\cite{obj_detection_survey_2} is deployed to extract small feature maps and keep the important local information. Then, temporal convolutions are applied to enhance the communication among time steps and fuse high-dimensional spatio-temporal features into low-dimensional temporal ones. 
This experiment shows that compared to global average pooling, STFM results in improved WER on LRW, LRS2-BBC, and LRS3-BBC.

\paratitle{\textbf{Power of Visemic Features.}}

C2D stresses visemic features but C3D encodes both spatial features and short-term (local) temporal dynamics in a sequence. Retrospective works show that using them jointly results in more powerful and discriminative features.
A common approach to combine them is to first give the input sequence to a C3D, and then to a deeper 2D network such as ResNet, extracting more fine-grained features~\cite{11pdf, PCPG}.
In this setting, the output of the C3D module, a tensor of size $T \times W \times H \times C$, can be divided into $\mathbf{t}$ various time steps and then each of them is fed to the C2D. It can ingests the whole C3D output tensor as a single time step as well~\cite{HPC,18pdf, 7pdf, 11pdf, 15pdf,12pdf,2pdf, 13pdf, MIM, teacher_student,26pdf,37pdf}.

Following that, ~\cite{HPC} use similar strategy, however, they employ Hierarchical Pyramidal Convolution~(HPC) to increase the contribution of 2D network. An HPC block is a pyramid with $\mathbf{n}$ levels of kernels with different sizes~($\mathbf{n}=4$ and $\mathbf{kernel\_size}= 3,5,7,9$ see~\cite{HPC} for more details). The small kernels and large ones are responsible to extract feature maps with local~(fine-grained details) and global contextual information, respectively. 
Moreover, there is hierarchical connection between adjacent layers of the pyramid to help information flow between different levels so that the local feature map is a part of the output. Using this bottom-up information aggregation, the model performance is improved, specially on words with few visemes. The final C2D in visual feature extractor is a HP-ResNet-$18$ which is similar to ResNet-$18$ but the second standard convolution layer of each basic block is replaced with an HPC block. 
The final VSR pipeline has better word accuracy on the LRW compared to D3D, inflated 3D, 3D convolution combined with baseline ResNet-$18$ and ResNet-$34$, and P3D-ResNet-$50$.

\paratitle{\textbf{Knowledge Distillation in Lip Reading.}}

Knowledge Distillation (KD)~\cite{Tang2020, Abbasi2020, Wu2019} is a common method to reduce the computational complexity and improve the inference latency without sacrificing too much performance. In this method, the knowledge of a deep, formidable model with high number of parameters, is transferred to a compact and less complex one. The first model, called teacher, has richer knowledge due to its higher learning capacity compared to the second one, called student. 
KD, as a model compression technique, is used in various fields, such as video action recognition~\cite{Wu2019} and image classification~\cite{Tang2020, Abbasi2020}. However, by introducing `born again neural networks'~(BANNs), ~\cite{bornAgain} demonstrated that KD can also be used to transfer knowledge from teachers to students with identical learning capacities. In context of lip reading, this idea has been used for isolated word recognition by ~\cite{teacher_student} to offer an extra supervisory signal with inter-class similarity information. The development procedure involves training a sequence of teacher-student classifiers in which the teacher of each step~(except the first one) is replaced with the student of the previous step until no improvement achieved. The trained models on LRW and LRW-1000 achieved better accuracy compared to those trained without distillation.

Lip reading datasets are orders of magnitude smaller than their audio only counterparts used
for development of ASR models. Besides all the methods introduced so far as solutions to overcome this issue, another
remedy is cross-modal distillation, that is, deployment of teacher-student manner to transfer knowledge of a network
trained on one modality to another network accepting different modality~\cite{li2019improving, afouras2020asr, zhao2020hearing, ren2021learning}. The rational of this method is the fact that acoustic signal contains complementary information to VSR, specifically for characters with subtle lip movements and different phonemes with almost identical visemes. 
Using this strategy,~\cite{li2019improving} proposed a network for Audio-Visual Speech Recognition (AVSR), comprising an audio-only teacher and an audio-visual student. The training procedure includes two main steps; first, the audio-based teacher is trained on an external audio data along with the audio data of an audio-visual dataset. In the second step, the audio-visual student is trained to minimize the Kullback-Leibler (KL) divergence between the student’s output and the  posteriors  generated  by the acoustic teacher.
This technique not only results in lower error rate but also with slight modification, it contributes to employ large scale unlabelled datasets to boost the VSR performance. ~\cite{afouras2020asr} exploited an audio-based pre-trained network as an ASR teacher and to generate transcripts for unlabelled videos. The pretrained model is then finalized on VSR specific datasets.

Aside from these gains, transferring knowledge from audio to video, two heterogeneous modalities, faces a critical concern: asynchronicity or various sampling rates of audio and video signals~\cite{jaimes2007multimodal}. This may occur due to different sampling rates of video and audio sequences, resulting in length inconsistency, and blanks at the beginning or end of the sequence. 
~\cite{zhao2020hearing} proposed a network that uses cross-modal alignment strategy to synchronize audio and video data by finding the correspondence between them.
In order to do so, frame-level KD helps to learn the alignment between audio and video based on the RNN hidden state vectors of the audio encoder and video encoder. More specifically, the most similar video frame feature is calculated by a way similar to the attention mechanism. 
Furthermore, to improve the performance of the final student network, the proposed model named Lip by Speech (LIBS), uses multi-granularity knowledge from speech recognizer: frame-, sequence-, and context-level.
At the first level of KD, frame-level distillation enables the network to learn more discriminative visual features by directly matching the frame-based visual feature with the corresponding audio one. Since both audio and video signals are different expressions of the same input, in the next step, sequence-level distillation tries to achieve similar video and audio feature vectors. Finally, to force the visual student to predict the same character as acoustic teacher, for each time step, context-level distillation push the corresponding context vectors to be the same.

Apart from the improvements brought by cross-modal distillation, more comprehensive examination shows that using acoustic teacher does not necessarily result in more accurate visual student~\cite{ren2021learning}. In fact, due to cross-modal gap, the teacher trained on video modality is a better supervisor for the visual student to learn more distilled knowledge representation. To address this issue, instead of using an audio-based teacher,~\cite{ren2021learning} developed a powerful `master' that is trained on both video and audio signals, producing three types of probabilities based on audio modality, video modality, and their combination. 
This design makes it possible for the student to make its own trade-offs on which modality to learn more from. Two separate `tutor' networks each pre-trained separately on audio and video modalities are used for this dynamic knowledge fusion. They generate fixed features that are encoded into weighting factors, measuring the contribution of audio and video modalities.
The experimental results show that using this strategy, the final student network has better WER compared to the networks trained only by audio supervision.

\paratitle{\textbf{Multi-Pathway Networks.}}

In a Multi-Pathway Network, it is common to feed various interconnected inputs to the same network. For example in video classification, a good example would be to feed the optical flow and raw frames of a video to the same network.

Early attempts of deploying such architectural design for lip reading was performed by ~\cite{16pdf} to tackle the problem of multi-view lip reading. In the proposed model, different views of the same utterance are fed to the same feature extractor and then the output vectors are concatenated for further processing.
The proposed model receives three streams (frontal, profile, and $45^\circ$ views) and is trained on a OuluVS2 in which all the angles are static and known in advance. The overall experiment demonstrates that the combination of different views, specifically (frontal and profile) and (frontal, profile, and $45^\circ$), improves the accuracy of frontal lip reading. 

Despite these results, the diversity of head poses during speaking makes this approach quite infeasible for other datasets specifically in the wild scenarios. This is due to the fact that the head pose angle of speakers is not annotated and if it would the quantity of these angles will result in a model with tens of streams. Thus, for lip reading application, probably a simpler approach to employ multi-pathway networks is to accepts different type of visual inputs, such as spatial gradient, or optical flow descriptors which are common options in video classification~\cite{Tang2015, Tang2019, 40pdf}.
Following that ~\cite{13pdf} proposed Deformation Flow Network (DFN) for word-level lip reading. Here, one modality is gray-scale video fed to a front-end module comprising C3D and ResNet-$18$. The other is deformation flow fed to another front-end encoder including C2D and ResNet-$18$. This flow is a mapping of the correlated pixels from the source one to the target frame and is generated by an endcoder-decoder trained in a self-supervised manner. By providing such mapping, DFN takes advantage of the subtle facial motion of consecutive frames for word-level prediction. 
Although, each branch predicts the word probabilities independently, they exchange information during training utilizing a bidirectional KD loss.

Another approach for input processing in multi-pathway approach is to feed the input to different neural networks each designed to capture various types of features and information~\cite{Sun2017, Bai2017}. In video classification field, SlowFast~\cite{slow_fast} is a two-pathway C3D that models information with different temporal resolution and provides complementary information for video classification. Wiriyathammabhum~\cite{SpotFAst} developed SpotFast network which is similar to SlowFast for word-level lip reading and achieved improved performance on LRW.
A SlowFast network has two pathways: `slow' and `fast';
The `slow' pathway \footnote{In the original paper~\cite{SpotFAst}, it is referred as `spot' pathway.} is designed to capture semantic information of consecutive frames, and it operates at low frame rates. On the other hand, the `fast' pathway is responsible for modeling abrupt motion changes, by operating at fast refreshing speed and high temporal resolution. The fast pathway is made very lightweight, since it has fewer channels and weaker ability to process spatial information. To fuse different levels of temporal resolution and semantic information, lateral connections are leveraged so that each pathway is aware of the representation learned by the other one.
This kind of network is suitable for the task of lip reading since the slow pathway can model the changes of lips when uttering a specific character and the fast pathway captures the lips motion when uttering a specific word.

\paratitle{\textbf{Target Word Boundaries.}}

For most of controlled lip reading datasets, each sample video only contains frames corresponding to the spoken token (i.e. character, word, sentence, etc.), and consequently the VSR models receiving well-annotated data, achieve high accuracy. On the other hand, in lip reading datasets in the wild, the target token is usually surrounded by other tokens as well, similar to the LRW sample labels mentioned in section ~\ref{sec:wilddatasets}. This characteristic makes the training procedure challenging, since the model not only requires to correctly classify the samples, but also to spot the target token and learn identical patterns.

In LRW, the target word happens approximately at the middle of the sample video containing 29 frames, so that the target word boundaries are  determined to some degree.
To use this characteristic and address issues associated with lack of exact word boundaries, ~\cite{stafylakis2018pushing} benefited from the start and end of target word annotation in LRW in two avenues; In the first approach, those frames not related to the target word utterance are removed. In the second one, the word boundaries are passed to the model as additional binary indicator features specifying whether or not the frame lies inside or outside the word boundaries.
The final results demonstrate the superiority of binary indicator variables compared to non-informative frames elimination.

In similar fashion,~\cite{feng2020learn} also confirmed these results and showed that the out-of-boundaries frames can provide contextual and environmental information (i.e. the speaker, pose, light, etc.) that is useful to distinguish the target word.

While binary indicator features yields substantial improvement, this type of information requires hard work of annotating the samples. Thus, a more dynamic approach is beneficial to make the model itself responsible to determine these boundaries. Moreover, as mentioned in section~\ref{sec:wilddatasets}, intrinsic characteristics of lip reading datasets in the wild, such as homophones, class agnostic variations (e.g. speaker head orientation and various lighting conditions), render the samples of each class nonhomogeneous.

To address these challenges, ~\cite{MIM} applied Mutual Information Maximization (MIM) constraint at local and global levels. Local Mutual Information Maximization (LMIM) helps to extract features representing word-related fine-grained movements at time step $t$. These features can help in homophone classification and to discriminate among different classes too.
On the other hand, Global Mutual Information Maximization (GMIM) extracts the mutual pattern of the same word in various videos and helps the model to locate the key frames representing the target word. 
Following this idea, in the front-end of the proposed pipeline, the video goes through 3D convolutions and ResNet-$18$ and the extracted features are then divided into $T$ time steps. The pairs of each time step features and the sample label are then fed to the LMIM module. 
GMIM module likewise ingests the features extracted by the front-end but captures the global information using Bi-GRUs and an LSTM and assigns different weights $\beta$ ($T \times 1$-dimensional) for different frames according to the target word. The experimental results validate the efficiency of both LMIM and GMIM on two word-level datasets, LRW and LRW-1000.

\paratitle{\textbf{Reinforcement Learning.}}

In a Sequence to Sequence (Seq2Seq) approach, as a common method for sequence modeling, the output in each time step tightly depends on the ground truth label of the previous one. This dependency leads to faster model convergence, but in the actual test process, no ground truth label is available and consequently if the model outputs a wrong label in time-step $t$, all the other predictions after that will be affected, i.e., the error will accumulate along the output sequence.~\cite{PCPG} tried to address this problem by developing a pseudo-convolutional policy gradient (PCPG) method for both word and sentence level lip reading. In this method, they also tried to answer the problem arises by the inconsistency between the optimized discriminative target (cross entropy) and the final non-differentiable evaluation metric (WER/CER). PCPG applies Reinforcement Learning (RL) into the Seq2Seq model to connect the optimized discriminative target and the evaluation metric directly. This model consists of a video encoder for spatio-temporal feature extraction and a GRU-based decoder to generate predictions at each time step by reward maximization. 

In the encoder block, the short-term and long-term dependencies between time steps are extracted by the 3D convolutions and ResNet-$18$, and Bi-GRU, respectively.
On the PCPG's decoder, a $2$-layer GRU is followed to decode each character at each output’s time step. In the learning process of PCPG, the optimization objective is to minimize the cross-entropy loss at each time step for which the output is decided by the predictions at the previous time steps. Moreover, Seq2Seq model is considered as an `agent' interacting with an external `environment' corresponding to video frames, words, or sentences here. In this way, the model can be viewed as a policy leading to an `action' of choosing a character to output.

\subsubsection{Sequential Feature Extractors}
In the process of lip reading, we pay attention not only to the shape of the speaker's lips but also to the lip's motion and the sequential connection among visemic features. Thus in a VSR pipeline, there must be a module to capture dynamics of lips in the frames. Sequential feature extractors reviewed in this work fall into $5$ categories: Classical techniques, C3D, Recurrent Neural Networks (RNNs), Temporal Convolutions (TCs), and Transformers.

\begin{figure*}[h]
    \centering
    \includegraphics[width=\textwidth]{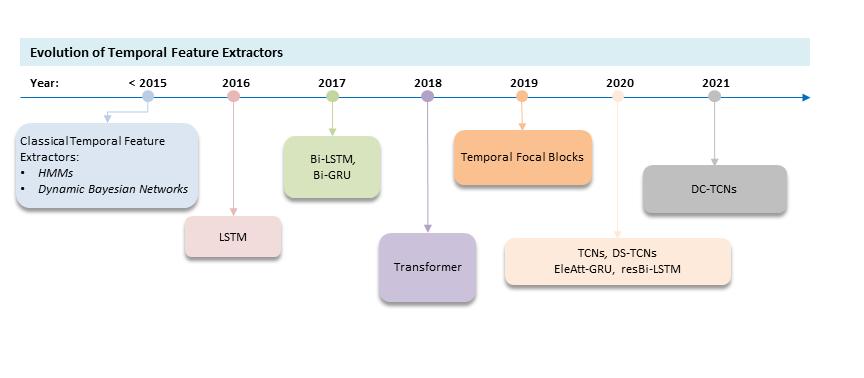}
    \caption{Evolution of Sequential Feature Extractors in Lip Reading Pipeline from 2015 to 2021.}
    \label{fig:temporalfeatureExctractor}
\end{figure*}

In section \ref{sec:visualfeatureextractor}, we introduced C3D as a spatial feature extractor capable of capturing short-term temporal dependencies, simultaneously. Thus in the rest of this section, we review the other methods and their characteristics. Figure \ref{fig:temporalfeatureExctractor} represents the evolution history of sequential modeling techniques in lip reading. 

\paragraph{Classical Methods}

Traditional lip reading systems mainly process handcrafted visual features using Hidden Markov Models (HMMs)~\cite{Zhou2014, LipReadingSurvey, 28pdf, delakis2008audiovisual, wu2016novel, dupont2000audio} and Dynamic Bayesian Networks (DBNs)~\cite{Saenko2009}. HMMs utilize short context information to characterize the dynamics in the feature space. 
DBNs, as a generalization of HMMs, also model temporal dependencies of lips' visual features. DBNs are developed to ameliorate high sample complexity and computational complexity of HMMs~\cite{Murphy2002}. 
According to the purpose of this work, we skip scrutinizing these methods but we refer the readers to~\cite{Zhou2014}, ~\cite{LipReadingSurvey}, and ~\cite{hao2020survey} for more information.

\paragraph{Recurrent Neural Networks}

RNNs are well-known in applications where there exist temporal dependencies among units of the input, such as language modelling, machine translation, speech recognition, and image captioning. 
In this type of networks, the hidden states acts as a representation of previously seen information and consequently the current output depends on both current input and the already processed outputs~\cite{time.pdf}. With this characteristic, RNNs are powerful enough to maintain long-term interrelations~\cite{HimansuDas2020}. 

Similar to the other deep models, RNNs suffers from the vanishing gradient problem and several gate-based RNN structures, such as LSTMs and GRUs, are proposed to tackle this problem~\cite{ArunKumarSangaiah2019}. Another variation of these networks is Bidirectional RNNs (Bi-RNNs). They attempt to exploit future events as well as previously seen information to determine the output. Since the words in a sentence can be logically related to either previous or subsequent words, Bi-RNNs can get both forward and backward information within the sequence~\cite{26pdf}.

Bidirectional and unidirectional RNNs are commonly used in lip reading pipelines~\cite{7pdf, LRWS, LipNet, 11pdf, 15pdf, 1pdf, LCANET, D3D, 13pdf, MIM, P3D, PCPG, 5pdf, 26pdf, 17pdf, Sterpu2018, CMLR, zhao2020hearing}, but it is worth noting that in practice, they may fail to learn more complicated information. As a result, various modifications have been made to improve their learning capacity. 
For example stacking several RNN layers is usual to model intricate patterns of the input sequence. Inspired by this structure,~\cite{resBi-LSTM} developed resBi-LTSM for speech recognition. To increase the learning capacity, they added residual connections to the Bi-LTSM that adds up the original features extracted by the CNN module to the output of the embedded RNN. As a result, with the help of new blocks, the phoneme information is passed to the deeper layers. The resBi-LTSM architecture is also used in the VSR method proposed by~\cite{26pdf}.
In the proposed pipeline, the features extracted by C3D and DenseNet model are processed by a two-layer resBi-LSTM. This experiment emphasizes that by fusing visemic and semantic motion information, resBi-LSTM learns more complicated lip reading patterns and the fast flow path after the CNN layers results in less WER.

In another attempt to improve the performance of RNNs, ~\cite{EleAtt} tried to benefit from gate mechanisms along with attention. They proposed Element-wise-Attention Gate (EleAttG) proven to be effective for action recognition, sequential image classification, and gesture recognition. Attention mechanism is an acceptable approach to develop high performance models in sequence analysis tasks, such as machine translation~\cite{Luong2015}. It also helps the model to identify the key temporal stages and selectively focus on salient parts of the source sequence.
~\cite{P3D} used the EleAttG in their lip reading model to build character relationships within long sequences. In the proposed multi-modality model, the audio and video inputs are processed by two separate decoders and then the fusion of generated context vectors is fed to a one-layer EleAtt-GRU encoder. In this experiment, the word accuracy and WER have been improved in word-level and sentence-level lip reading, respectively.

\paragraph{Transformers}

RNN and its variations permits to avoid the need for aligning the symbol positions of the input and output sequences as this alignment can be calculated in computation time. But, this characteristic causes a sequential dependency between the hidden states making the parallelization during development phase unfeasible. This parallelization is necessary for long sequences and when there is memory limit to increase the batch size.
Transformers, on the other hand, avoid this recurrency by using self-attention mechanism to relate different positions of a single sequence~\cite{Attention}. Inevitably, fewer number of sequential operations leads to more parallelization. 
Unlike RNNs that receive each input at a time, transformers process the whole input sequence at once which makes them faster but results in loosing the critical information related to the ordering of the input sequence. The positional encoding mechanism is a solution to this problem and injects the ordering information into sequence processing procedure.

A basic Transformer is an encoder-decoder structure with multi-head attention layers, each focusing on different representation sub-spaces~\cite{Afouras2018a}. In the encoder, the input tensor which is served as the the attention query, key, and value, goes through a stack of self-attention layers.
But, every decoder ingests the encoder's output as the attention key and value and the previous decoding layer's output  as the  query. The ordering information is fed to the model via fixed positional embeddings in the form of sinusoid functions.  The decoder outputs the character probabilities trained with a cross-entropy loss. The Transformers implicitly learn a language model during training, thus there is generally no need for an explicit one, although experiments show that it could be beneficial~\cite{Kannan2017}.

The basic structure of Transformers demonstrate decent performance on lip reading~\cite{12pdf, SpotFAst, 37pdf}.~\cite{12pdf} developed two Audio-Visual Recognition (AVR) models based on the Transformers self-attention architecture (TM). One is trained with Seq2Seq loss (TM-Seq2Seq) and the other with Connectionist Temporal Classification (CTC) loss (TM-CTC) for sentence level lip reading. In TM-Seq2Seq, each modality is processed by separate attention heads and then the concatenation of the resulting video and audio context vectors is propagated to the feedforward block. In contrast, in TM-CTC, the concatenation of the audio and video encodings is propagated through the stack of self-attention and feedforward blocks.
The results show that TM-Seq2Seq achieved better WER for the visual modality and even using an external language model with appropriate beam width yields over $22\%$ WER reduction compared to the previously SOTA model on the same dataset. On the other hand, both TM-Seq2Seq and TM-CTC achieved the same gain when using both audio and visual modalities. In the proposed AVSR systems, the visual cues offered by the mouth frames gives an improvement not only when the audio signal is noisy but also when it is clean. 

The authors also assessed the performance of the AVSR models for out-of-sync audio and video inputs. They synthetically shifted the video frames to achieve an out-of-sync inputs. The evaluation process showed that the TM-Seq2Seq architecture is more resistant to these shifts and even without fine-tuning on out-of-sync data, its performance is superior to the TM-CTC counterpart. This result emphasizes the advantage of independent encoder-decoder attention mechanisms for each modalities.

\paragraph{Temporal Convolutions}
In context of sequence modeling, RNNs are among well-adopted solutions. However, for long input sequences, LSTMs and GRUs require a lot of memory to store the partial results for their multiple cell gates and the training procedure is notoriously difficult~\cite{time.pdf}. The other substitute, Transformer, also suffers from complex and time consuming training procedure. Furthermore, it builds character relationships within limited length and is less effective with long sequences as compared to RNNs~\cite{P3D}. 

Another sequential feature extractor is Temporal Convolutional Network (TCN) architecture. Recent research indicates that it can outperform baseline RNNs in various tasks including but not limited to ASR, word-level language modeling, and machine translation~\cite{bai2018empirical}.

As stated earlier, the main characteristic of RNNs is capturing long dependencies in sequences, nevertheless, a comparison study demonstrates that TCNs hold stronger memory retention as compared to RNNs and consequently are more suitable for domains where a long history is required~\cite{time.pdf}.
Further, TCNs exploit 1D fully-convolutional network (FCN); allowing to produce output tensor with the same length of the input tensor.

In context of lip reading, various modifications have been applied on TCNs to make them more appropriate for the application. For instance, to enable TCN capture more complex temporal patterns, multi-scale TCN benefits from combining feature vectors extracted by various kernel sizes~\cite{18pdf}. It processes the input in multiple temporal scales and combines both short term and long term information~\cite{18pdf}.
This TCN variant is employed in the lip reading model proposed by ~\cite{18pdf} in which, the video frames are fed to a multi-scale TCN, after being processed by the C3D and ResNet-$18$ blocks. The proposed model was used for word-level lip reading and outperforms the RNN-based counterparts by a considerable margin.
~\cite{TCN_2} also developed a word-level VSR system benefiting from  the multi-scale TCN (Densely Connected Temporal Convolutional Network (DC-TCN)) along with channel-wise attention layer, covering temporal scales in a denser fashion. They also applied various temporal kernel sizes in a sequential manner which was less effective than multi-scale TCN. 

Multi-scale TCNs are proven to be effective and to extract more robust temporal features, but their computational costs are non-negligible. In an attempt to reduce the model's computational complexity, development of Depth-wise Separable TCN (DS-TCN) results in a temporal feature extractor with fewer FLOPs~\cite{teacher_student}.

Temporal Focal block is another type of sequential feature extractor based on 1D convolutions (C1D)~\cite{24pdf}. The simple implementation of TF-block consists of two convolutional layers, each followed by layer normalization and Relu activation. In this block, using different kernel sizes help to extract correct semantic information and to learn more robust representations, as well. 
Using TF-blocks,~\cite{24pdf} developed a fully convolutional VSR pipeline which is more time and memory efficient compared to Transformers with slow training speed-- a feature that greatly limits the models transfer learning ability. In the proposed conv-Seq2Seq model, the visual features are fed to an encoder-decoder structure for sequence modeling. In the encoder module, the position encodings, similar to those used in Transformers, are added to the features at the bottom of the encoder to model ordering information of the sequence. Moreover, TF-blocks and self-attention mechanism are used to capture short-range and long-range temporal dependencies, respectively.

In the decoder module, the previously predicted character embedding goes through a multi-head attention module and its output along with the encoder's output are then fed to a multi-head vanilla attention module\footnote{The attention weights are derived from the decoder hidden states and all the encoder output states}. Moreover, the decoder should be future-blind so that causal convolution is used in the decoder's TF-blocks. 

The proposed fully convolutional model outperforms the RNN-based opponents for both sentence and word level lip reading. Nevertheless, the Transformer-based models achieved better WER for sentence-level lip reading with the cost of time consuming training procedure.

As a complement to casual convolutions, we should emphasise that the training procedure is faster (no recurrent connection) and the model does not violate the ordering in which the data is processed~\cite{time.pdf, causal}. However, this type of convolutions limits the access to the history and requires many layers, or large filters to increase the receptive field~\cite{time.pdf}. In addition, in lip reading, the future-blind characteristic of causal convolutions is not really required. An alternative would be dilated temporal convolution (non-causal convolution) that have an exponentially large receptive field and include future information. Although, this variation of TCNs is fast to train as well as causal convolutions, it has never been employed in VSR pipelines. 

\subsection{Classification Module}

In the VSR pipeline, classification is the final step. The classification layer is a softmax layer providing probability distribution over classes. The output of this module can be character, viseme, phoneme, word, phrase, and sentence \footnote{Sentence level recognition is infrequent as compared to the rest.} (at prediction level). On the other hand, the largest spoken unit in the input video, can be word, phrase, or sentence (at recognition level). Thus, the system needs to make a connection between the prediction and recognition levels to compute the correct output. In this step, approaches tightly depends on the amount of information fed to the softmax layer and generally fall into three categories: (i) Direct-softmax, (ii) Seq2Seq, and (iii) CTC.

In the following, we provide more details about each approach.

\paratitle{\textbf{Direct-softmax}}:
An intuitive approach is to consider recognition and prediction at the same level and to feed the final extracted features to the softmax layer, at once.
This method does not require any post-processing step, such as language models, and is a common choice when the largest spoken unit in the dataset is not sentence (i.e., character, word, or phrase)~\cite{LRW, D3D, 11pdf, 15pdf, 13pdf, MIM, SpotFAst, teacher_student, 5pdf, 18pdf, HPC, pose_invarient, TCN_2}.

\paratitle{\textbf{Sequence to Sequence}}: This approach is a popular method for both VSR and ASR~\cite{12pdf}.
In a Seq2Seq model, the tensor of extracted features is divided into equal time steps and then each of them is fed to the classification layer. Moreover, the output at time $t$ is conditioned on previous outputs, i.e., $1:t-1$ so that, the model implicitly learns a language model over output symbols and no further processing is required. Additionally, the model makes full use of global information of longer sequences~\cite{26pdf}.
Lip reading is usually considered as a Seq2Seq challenge and a good number of the proposed methods falls into this category~\cite{12pdf, LRWS, 24pdf, P3D, PCPG, 37pdf, 17pdf, Sterpu2018}.

\paratitle{\textbf{CTC}}: In character-level prediction, the training sequence must be aligned to the output labels (a character for each time step), even though, the input and target sequences are not generally the same length.
This alignment can be done by human experts but it is cumbersome for large datasets. 
A solution to mitigate this problem is CTC loss function~\cite{CTC}; eliminating the need for prior alignment between the input and output sequences.
Using this loss function, the labels are predicted for each time step (e.g. frame-wise) in isolation with others. This manner is a potential weakness that can be alleviated by a language model employed as a post-processing step. 
When using CTC loss, the vocabulary, which is the set of tokens that the model predicts, includes a `blank' character denoting as `-' that helps to encode duplicate characters.For instance, in CTC configuration, the ground truth form of word `Hello' is `Hel-lo', which means that the character `l' is repeated twice.
CTC loss functions works intuitively; It receives the model's output matrix containing a score for each token at each time-step and the ground truth sequence.
In training phase, the objective will be to maximize the probability of paths leading to the ground truth label or to minimize the negative sum of log-probabilities. In the validation and test phase, for each time step, using a beam search or greedy approach, a character is selected and after deleting the repeated characters and blanks, the remaining sequence is the final recognition output.
It worth noting that, the CTC loss function can be applied on character, viseme, or phoneme levels. 

At the end of this section, we should also mention that when the prediction level is phoneme, viseme, or Hanyu Pinyin, there must be a module to provide a mapping between the character and the softmax's  output. For example, in ~\cite{26pdf}, the softmax layer outputs the Hanyu Pinyin probabilities, thus, there is another methods to map the Hanyu Pinyin to Chinese characters. 

\section {Promising Future Directions and Opportunities}\label{sec:future}

Having discussed key advances and challenges, we now envision some promising future directions and concerns.

\textbf{Lightweight and Fast VSR}: 
In recent years with the development of smart phones, the popularity of Speech-to-Text applications have been increased and the fusion of speech modality and lip movements has led to more robust speech recognition in real-world applications. For instance, Liopa~\footnote{\url{https://liopa.ai/.}} is a mobile application with such goal that also provides voiceless speech recognition and silent communication. 

Despite being accurate, the networks used in these applications need to be fast and light weight, but most of them have millions to hundreds of millions parameters making them unsuitable for mobile devices. The development and test phase of such deep CNNs used for feature extraction also require high computational resources such as GPUs\footnote{Graphics Processing Units}. Furthermore, Transformers and RNNs, which usually are an inseparable part of most VSR systems, have proven to be computational intensive~\cite{kouris2020approximate}.  Both spacial and sequential feature extractors suffer from not only high inference time but also time consuming training procedure.
The extra time for input preparation is also non negligible. Moreover, in test phase, the input video usually includes the spoken unit and several silent frames at the start and end of the video. To specify the boundaries of the utterance and feed the exact amount of information to the model, a lip activity detection method is required, adding up further time to the input preparation step.

Probably, using compact and lightweight networks (i.e. MobileNet~\cite{Tian} and ShuffleNet~\cite{Shufflenet}) and network acceleration techniques (such as Network Pruning and Quantification~\cite{obj_detection_survey_2}) could be the future of lip reading systems, especially for practical applications. Moreover, KD can be a proper solution to transfer the knowledge of large accurate models to light networks suitable for on-device applications. 
Moreover, using real-time and light-weight face detection/landmark prediction methods and temporal convolutions can also improve the input preparation and inference latency, respectively.

\textbf{Weakly Supervised Lip Reading}: As we mentioned before, one of the main obstacles to achieve effectiveness in lip reading, is the lack of large amount of well-annotated videos. The ultimate goal in lip reading field is to develop a VSR model capable of accurately and efficiently deciphering unconstrained natural language sentences uttered in videos in the wild. Current lip reading datasets contain only a few dozen to hundreds of categories, significantly fewer than those which can be recognized by human. Thus, new large-scale datasets with significant vocabulary and utterance sizes are required. But the annotation process is time-consuming, expensive, and inefficient.

Clean labeled data is a real concern for any supervised learning method. In context of image classification, a common approach is to use weak supervision technique~\cite{Tian}. 
This technique has never been employed for lip reading projects but it can reduce human labor costs in video annotation process.
On the other hand, few-shot and zero-shot learning methods are also very appealing specifically if we consider lip reading as an `open-world' problem~\cite{fewshotlearning, zeroshotlearning}.

\textbf{Pre-Training and Fine Tuning in Lip Reading}:
In the training procedure of a CNN, the model weights are initialized by values randomly sampled from a normal distribution with zero mean and small standard deviation~\cite{Tajbakhsh2016}. In applications like lip reading, where there is a large number of weights in the CNN and restricted access to the labeled data, this may yield to an undesirable local minimum for the cost function.
Alternatively, we may set the weights of the convolutional layers to those of another model with the same architecture trained on different dataset. This pre-trained model improves generalization ability and convergence speed of the final model. In addition, it has already learned to extract lip features, so that in the second training round, the current visemic specification of the current dataset will be learned.~\cite{jitaru2021toward} demonstrated the effectiveness of this approach, although further examination is required.

\section{Conclusions}\label{sec:conclusion}
For a long period of time, handcrafted visual and temporal features in traditional lip reading systems failed to model the crucial details of lips movements and changes when uttering a specific word. Consequently, due to its limited effectiveness, researchers only considered visual clues as a complementary information for potential applications such as speech to text. However, recent remarkable performance of deep models processing solely visual modality have validated that, as an independent approach, VSR is a practical solution to a variety of other applications such as visual passwords and law enforcement. On the other hand, in contrast to other video and image related tasks, lack of precisely labeled and large-scale datasets was another impediment to the progress of lip reading methods. This hindrance also has been alleviated by developing a great number of audio-visual speech datasets.
These advancements achieved by deep models and high quality data resulted in numerous efforts to design and develop accurate lip reading methods.
In this survey, we thoroughly investigated those efforts, with the aim to summarize the existing studies and to provide insightful discussions. 

To better understand how a VSR system works, we divided the basic building blocks of a pipeline in three sub-modules: input preparation, feature extraction, and classification. We discussed the purpose of each module, the most controversial and task-specific challenges, and how the retrospective works faced them. Additionally, we categorized the most popular lipreading datasets according to the recording settings and extensively discussed the details and related data complications.
Furthermore, we also provided some insights for future research directions and have listed the still open issues. We hope this survey helps researchers to develop novel ideas with new perspective.
Attributed to its distinct characteristics, the research field of lip reading is still far from complete and its performance is not as high as the other computer vision applications such as object detection, or video/image classification. However, we believe this paper will help readers to build a big picture and to find future directions of this fast-moving research field.


\begin{landscape}
\begin{table}[]
\tiny
\centering
\caption{The Statistics of Lip Reading datasets (M: Male, F: Female); (**: Only visual modality is available).}
\label{tab:table1}
\begin{tabular}{@{}ccccp{20mm}ccccccc@{}}
\toprule
\textbf{Data Set   Name} &
  \textbf{Language} &
  \textbf{Classification Level} &
  \textbf{Recording Setting} &
  \textbf{Available} &
  \textbf{\# Utterance} &
  \textbf{Speakers} &
  \textbf{Vocab size} &
  \textbf{Year} &
  \textbf{} &
  \textbf{} \\ \midrule
AVLetters~\cite{AVLetter} &
  English &
  Isolated letters &
  Controlled &
  Available by contact &
  780 &
  10 (5M/5F) &
  26 &
  2002 &
   &
   \\
Tulips1**~\cite{Tulips1} &
  English &
  Digits &
  Controlled &
  * &
  96 &
  12 (9M/3F) &
  4 &
  1995 &
   &
   \\
DAVID~\cite{DAVID} &
  English &
  Digits/Sequence of letters &
  Controlled &
  * &
  178 &
  \begin{tabular}[c]{@{}c@{}}124(64M/\\      61F)\end{tabular} &
  * &
  1996 &
   &
   \\
AVICAR~\cite{AVICAR} &
  English &
  Digits/Letters/Sentence &
  Car &
  Public audio, but \newline the video is only available by contact &
  59k &
  100 (50F/50M) &
  * &
  2004 &
   &
   \\
CUAVE~\cite{CUAVE, CUAVE_2} &
  English &
  Isolated/Continuous digits &
  Controlled &
  Available by contact &
  6960 &
  36 (19M/17F) &
  * &
  2002 &
   &
   \\
OuluVS2~\cite{8pdf} &
  English &
  Continuous digits/Phrase/Sentence &
  Controlled &
  Available by contact &
  20k &
  53(40M/13F) &
  * &
  2015 &
   &
   \\
LRW~\cite{LRW}&
  English &
  Word &
  Wild &
  Available by contact &
  539k+ &
  * &
  500 &
  2016 &
   &
   \\
LRW-1000~\cite{D3D} &
  Mandarin &
  Word &
  Wild &
  Available by contact &
  718,018 &
  2,000 &
  1000 &
  2018 &
   &
   \\
Greek-words~\cite{Greek} &
  Greek &
  Word &
  Controlled &
  Public &
  2500 &
  10(6M/4F) &
  50 &
  2019 &
   &
   \\
Wild LRRo~\cite{LRRo} &
  Romanian &
  Word &
  Wild &
  * &
  1087 &
  \textgreater{}35(64-66\% M) &
  21 &
  2019 &
   &
   \\
Lab LRRo~\cite{LRRo} &
  Romanian &
  Word &
  Controlled &
  * &
  8180 &
  19 &
  48 &
  2020 &
   &
   \\
LSVSR~\cite{1pdf}&
  English &
  Sentence &
  Wild &
  * &
  2,934,899 &
  * &
  127,055 &
  2018 &
   &
   \\
MIRACL-VC**~\cite{MIRACL} &
  English &
  Word/Phrase &
  Controlled &
  Public &
  3k &
  15(5M/10F) &
  * &
  2014 &
   &
   \\
OuluVS~\cite{OuluVS} &
  English &
  Phrase &
  Controlled &
  Available by contact &
  817 &
  20(17M/3F) &
  * &
  2009 &
   &
   \\
LRS~\cite{LRWS} &
  English &
  Sentence/Phrase &
  Wild &
  Available by contact &
  118k+ &
  * &
  17428(train/val), 6,882(test) &
  2016 &
   &
   \\
LRS2-BBC~\cite{12pdf} &
  English &
  Sentence/Phrase &
  Wild &
  Available by contact &
  144k+ &
  * &
  41k(pre-train), 18k(train/val),   1,693(test) &
  2018 &
   &
   \\
LRS3-TED~\cite{LRS3} &
  English &
  Sentence/Phrase &
  Wild &
  Available by contact &
  152k+ &
  9545 &
  52k(pre-train), 17k(train/val),   2136(test) &
  2018 &
   &
   \\
TCD-TIMIT~\cite{TCD} &
  English &
  Sentence &
  Controlled &
  Public &
  13826 &
  62(32M/30F) &
  98 sentences for volunteers; 377   sentences for lipspeakers &
  2915 &
   &
   \\
MV-LRS~\cite{17pdf} &
  English &
  Sentence &
  Wild &
  Available by contact &
  500k+ &
  * &
  30k(pre-train), 15k(train/val),   4311(test) &
  2017 &
   &
   \\
NSTDB~\cite{26pdf} &
  Mandarin &
  Sentence &
  Wild &
  * &
  1,705 &
  * &
  349 &
  2020 &
   &
   \\
CMLR~\cite{CMLR} &
  Mandarin &
  Sentence &
  Wild &
  Public &
  102,076 &
  * &
  * &
  2019 &
   &
   \\
VRLF~\cite{VRLF} &
  Spanish &
  Sentence &
  Controlled &
  Public &
  600 &
  24(3M/21F) &
  * &
  2017 &
   &
   \\
VidTIMIT~\cite{33pdf} &
  English &
  Sentence &
  Controlled &
  Available by contact &
  430 &
  43(24M/19F) &
  * &
  2009 &
   &
   \\
XM2VTS~\cite{XM2VTSDB} &
  English &
  Sentence &
  Controlled &
  Available by contact &
  1770 &
  295 &
  * &
  1999 &
   &
   \\
MODALITY~\cite{27pdf} &
  English &
  Word/Digit/Phrase &
  Controlled &
  Public &
  504 &
  35(26M/9F) &
  231 &
  2017 &
   &
   \\
SFAVD~\cite{SFAVD}&
  Farsi &
  Sentence &
  Controlled &
  Available by contact &
  587 &
  1M &
  * &
  2015 &
   &
   \\
CRSS-4ENGLISH-14~\cite{tao2018gating}&
  English &
  Word/Phrase/Continuous digits/Question/Sentence &
  Controlled &
  * &
  * &
  442 (225M/217F) &
  * &
  2018 &
   &
   \\ \bottomrule
\end{tabular}
\end{table}
\end{landscape}

\begin{landscape}
\begin{table}[]
\tiny
\centering
\caption{Deep Learning Based Lip Reading Systems(V: Visual, AV: Audio-Visual)}
\label{tab:table2}
\begin{tabular}{@{}cccccccccccc@{}}
\toprule
\textbf{Title} &
  \textbf{Visual Feature Extraction} &
  \textbf{Temporal Feature Extraction} &
  \textbf{Recognition/Prediction level} &
  \textbf{Modality} &
  \textbf{Dataset} &
  \textbf{WER(\%)} &
  \textbf{CER(\%)} &
  \textbf{Accuracy(\%)} &
  \textbf{BLEU} &
  \textbf{Year} \\ \midrule
  ~\cite{7pdf} &
  C2D/C3D &
  LSTM/C3D &
  Phrase/ Phrase &
  V &
  OuluVS2 &
  * &
  * &
  83.8 &
  * &
  2016 \\
  ~\cite{LRWS} &
  C2D &
  LSTM &
  Sentence/ Character &
  AV &
  LRS, LRW, GRID &
  \begin{tabular}[c]{@{}c@{}}(50.2, 23.8, 3(V)),\\  (13.9, *, *(AV-clean))\end{tabular} &
  \begin{tabular}[c]{@{}c@{}}(39.5, *, *(V))\\ (7.9, *, *(AV-clean))\end{tabular} &
  * &
  \begin{tabular}[c]{@{}c@{}}(54.5, *, *(V)), \\  (87.4, *, *(AV))\end{tabular} &
  2016 \\
  ~\cite{LRW} &
  C2D/C3D &
  C3D &
  Phrase, Word/ Phrase, Word &
  V &
  OuluVS, OuluVS2, LRW &
  * &
  * &
  91.4, 93.2, 61.1 &
  * &
  2017 \\
  ~\cite{LipNet} &
  C3D &
  BiLSTM/C3D &
  Sentence/ Character &
  V &
  GRID &
  \begin{tabular}[c]{@{}c@{}}11.4 (unseen speakers),\\  4.8 (seen   speakers)\end{tabular} &
  \begin{tabular}[c]{@{}c@{}}6.4 (unseen speakers), \\ 1.9 (seen   speakers)\end{tabular} &
  \begin{tabular}[c]{@{}c@{}}95.2   (unseen speakers),\\  86.4 (overlapped speakers)\end{tabular} &
  * &
  2017 \\
  ~\cite{11pdf} &
  C2D/C3D &
  LSTM/C3D &
  Word/ Word &
  V &
  LRW &
  * &
  * &
  83 &
  * &
  2017 \\
  ~\cite{17pdf} &
  C2D &
  LSTM &
  Phrase, Sentence/ Character &
  V &
  MV-LRS, OuluVS2 &
  62.8, * &
  54.4, * &
  *, 91.1(frontal) &
  42.5, * &
  2017 \\
  ~\cite{15pdf} &
  C2D/C3D &
  BiGRU/C3D &
  Word/ Word &
  AV &
  LRW &
  * &
  * &
  97.7(A), 82(V), 98(AV) &
  * &
  2018 \\
  ~\cite{12pdf} &
  C2D/C3D &
  Transformer &
  Sentence/ Character &
  AV &
  LRS2-BBC, LRS3-TED &
  \begin{tabular}[c]{@{}c@{}}(CTC: 54.7(V)), (S2S:48.3(V));\\  (CTC: 66.3(V)),   (S2S:58.9(V))\end{tabular} &
  * &
  * &
  * &
  2018 \\
  ~\cite{1pdf} &
  C3D &
  BiLSTM/C3D &
  Sentence/ Phoneme &
  V &
  LSVSR, LRS3-TED &
  40.2+-1.2, 55.1+-0.9 &
  28.3+-0.9 &
  * &
  * &
  2018 \\
  ~\cite{LCANET} &
  C3D &
  BiGRU/C3D &
  Sentence/ Character &
  V &
  GRID &
  3 &
  1.3 &
  * &
  * &
  2018 \\
~\cite{stafylakis2018pushing} &
  C3D &
  BiLSTM/C3D &
  Word/ Word &
  AV &
  LRW &
  11.92 &
  * &
  * &
  * &
  2018 \\
  ~\cite{D3D} &
  C3D &
  BiGRU/C3D &
  Word/ Word &
  V &
  LRW-1000, LRW &
  * &
  * &
  33, 78 &
  * &
  2019 \\
  ~\cite{24pdf} &
  C2D/C3D &
  C1D/C3D  &
  Sentence, Word/ Character &
  V &
  GRID, LRW, LRS2, LRS3 &
  1.3, 16.3, 51.7, 60.1 &
  * &
  * &
  * &
  2019 \\
  ~\cite{2pdf} &
  2D/C3D &
  C1D/C3D &
  Sentence/ Character &
  AV &
  LRS2 &
  5.93 &
  * &
  * &
  * &
  2019 \\
  ~\cite{CMLR} &
  C2D &
  BiGRU(encoder)/GRU(decoder) &
  Sentence/ Character &
  V &
  CMLR &
  * &
  32.48 &
  * &
  * &
  2019 \\
  ~\cite{zhao2020hearing} &
  C2D &
  BiGRU(encoder)/GRU(decoder) &
  Sentence/ Character &
  V &
  CMLR, LRS2 &
  *, 65.29 &
  31.27, 45.53 &
  * &
  69.99, 41.91 &
  2019 \\
  ~\cite{13pdf} &
  C2D/C3D &
  BiGRU/C3D &
  Word/ Word &
  V &
  LRW-1000, LRW &
  * &
  * &
  41.93, 84.13 &
  * &
  2020 \\
  ~\cite{MIM} &
  C2D/C3D &
  BiGRU/C3D &
  Word/ Word &
  V &
  LRW-1000, LRW &
  * &
  * &
  38.79, 84.41 &
  * &
  2020 \\
  ~\cite{P3D} &
  C3D &
  EleAtt-GRU/C1D/C3D &
  Sentence, Word/ Word &
  AV &
  LRW, LRS3-TED &
  20.7(LRS3-TED) &
  * &
  84.8(LRW) &
  * &
  2020 \\
  ~\cite{PCPG} &
  C2D/C3D &
  BiGRU/C3D &
  Sentence, Word/ Character &
  V &
  GRID, LRW, LRW-1000 &
  12.3, 22.7, 66.9 &
  5.9, 14.1, 51.3 &
  *, 83.5, 38.7 &
  * &
  2020 \\
  ~\cite{SpotFAst} &
  C3D &
  Transformers/C1D/C3D &
  Word/ Word &
  V &
  LRW &
  * &
  * &
  84.4 &
  * &
  2020 \\
  ~\cite{teacher_student} &
  C2D/C3D &
  C1D/C3D &
  Word/ Word &
  V &
  LRW, LRW-1000 &
  * &
  * &
  88.6, 46.6 &
  * &
  2020 \\
  ~\cite{5pdf} &
  C2D &
  BiLSTM &
  Phrase/ Phrase &
  V &
  OuluVS2 &
  * &
  * &
  * &
  * &
  2020 \\
  ~\cite{26pdf} &
  C2D/C3D &
  resBiLSTM/C3D &
  Sentence/ Hanyu Pinyin &
  V &
  NSTDB(Mandarin) &
  50.44 &
  * &
  * &
  * &
  2020 \\
  ~\cite{18pdf} &
  C2D/C3D &
  C1D/C3D &
  Word/ Word &
  V &
  LRW, LRW-1000 &
  * &
  * &
  85.3, 41.4 &
  * &
  2020 \\
  ~\cite{37pdf} &
  C2D/C3D &
  Transformer/C3D &
  Sentence/ Viseme &
  V &
  LRS2 &
  35.4 &
  23.1 &
  word accuracy(64.4) &
  * &
  2020 \\
  ~\cite{HPC} &
  C2D/C3D &
  C1D/C3D &
  Word/ Word &
  V &
  LRW &
  * &
  * &
  86.83 &
  * &
  2020 \\
  ~\cite{pose_invarient} &
  C2D/C3D &
  BiGRU/C3D &
  Word, Sentence/ Word &
  V &
  LRW, LRS2 &
  * &
  * &
  79.53, 59.60 &
  * &
  2020 \\
  ~\cite{afouras2020asr} &
  C2D/C3D &
  LSTM/C3D &
  Sentence/ Word &
  AV &
  LRS2, LRS3 &
  51.3, 59.8 &
  * &
  * &
  * &
  2020 \\
  ~\cite{feng2020learn} &
  C2D/C3D &
  BiGRU &
  Word/ Word &
  V &
  LRW, LRW-1000 &
  * &
  * &
  88.4, 55.7 &
  * &
  2020 \\
  ~\cite{TCN_2} &
  C2D/C3D &
  C1D /C3D &
  Word/ Word &
  V &
  LRW, LRW-1000 &
  * &
  * &
  88.36, 43.65 &
  * &
  2021 \\
  ~\cite{ren2021learning} &
  C2D/C3D &
  Transformer &
  Sentence, Word/ Word &
  AV &
  LRW, LRS2, LRS3 &
  14.3, 49.2, 59.0 &
  * &
  * &
  * &
  2021 \\\bottomrule
\end{tabular}
\end{table}
\end{landscape}